\documentclass[twoside]{article}

\usepackage[accepted]{aistats2024}

\usepackage[ruled,vlined]{algorithm2e}
\usepackage{subfigure}
\usepackage{multirow}
\usepackage{wrapfig}

\usepackage[utf8]{inputenc} 
\usepackage[T1]{fontenc}    
\usepackage{hyperref}       
\usepackage{url}            
\usepackage{booktabs}       
\usepackage{amsfonts}       
\usepackage{nicefrac}       
\usepackage{microtype}      
\usepackage{xcolor}         
\usepackage{graphicx}
\usepackage{amsmath, nccmath}
\usepackage{amsthm}
\usepackage{thmtools}
\usepackage{float}
\usepackage[english]{babel}
\usepackage{caption}
\usepackage[round]{natbib}
\newtheorem{lemma}{Lemma}

\begin{document}

\runningtitle{Online Calibrated and Conformal Prediction Improves Bayesian Optimization}

\twocolumn[

\aistatstitle{Online Calibrated and Conformal Prediction\\Improves Bayesian Optimization}

\aistatsauthor{ Shachi Deshpande \And  Charles Marx \And Volodymyr Kuleshov }

\aistatsaddress{Cornell Tech and Cornell University \And Stanford University \And Cornell Tech and Cornell University} ]



\begin{abstract}

Accurate uncertainty estimates are important in sequential model-based decision-making tasks such as Bayesian optimization. However, these estimates can be imperfect if the data violates assumptions made by the model (e.g., Gaussianity). This paper studies which uncertainties are needed in model-based decision-making and in Bayesian optimization, and argues that uncertainties can benefit from calibration---i.e., an 80\% predictive interval should contain the true outcome 80\% of the time.
Maintaining calibration, however, can be challenging when the data is non-stationary and depends on our actions.
We propose using simple algorithms based on online learning to provably maintain calibration on non-i.i.d.~data, and we show how to integrate these algorithms in Bayesian optimization with minimal overhead. 
Empirically, we find that calibrated Bayesian optimization converges to better optima in fewer steps, and we demonstrate improved performance on standard benchmark functions and hyperparameter optimization tasks.

\end{abstract}

\section{INTRODUCTION}

Bayesian optimization has emerged as a powerful tool for 
optimizing objective functions that are initially unknown and that are learned via evaluation queries
\citep{thornton2013autoweka,Shahriari2016BOReview,Bergstra_NIPS2011_HPOAlgs}.
%
In practice, querying such objectives can be expensive: for example, in hyperparameter optimization \citep{snoek2012practicalbo}, queries may involve training a machine learning model from scratch. 
Bayesian optimization aims to minimize the number of objective function queries by relying on a probabilistic model to guide search \citep{frazier2018tutorial}. 
However, probabilistic models are not always accurate and may be overconfident, which slows down optimization and may yield suboptimal local optima  \citep{guo2017calibration}.

%

This paper aims to improve the performance of algorithms for
online sequential decision-making under uncertainty, particularly Bayesian optimization. We start by examining which uncertainties are needed in model based decision-making 
and argue 
that ideal uncertainties should be calibrated \citep{gneiting2007strictly,kuleshov2018accurate}.
Intuitively, calibration means that an 80\% confidence interval should contain the true outcome 80\% of the time.
Calibration helps balance exploration and exploitation, estimate the expected value of the objective, and reach the optimum in fewer steps \citep{malik2019calibrated}.

Enforcing calibration in sequential tasks is challenging because the data is non-stationary (it is determined by our actions). 
We introduce simple algorithms based on online learning \citep{shalev2012online} that provably yield calibrated uncertainties, even on data chosen by an adversary.
Moreover, our methods can be added to any Bayesian optimization algorithm with minimal modifications and with minimal computational and implementation overhead.
Empirically, we show that our techniques improve Bayesian optimization and yield faster convergence to higher quality optima across a range of standard benchmark functions and hyperparameter optimization tasks.

\textbf{Contributions} In summary, this work makes the following contributions: (1) we study which uncertainties are needed in online sequential decision-making and and provide a formal analysis of the benefits of calibrated uncertainties; (2) we introduce simple algorithms than enforce calibration on non-stationary data and that can be added within any existing Bayesian optimization algorithm with minimal computational and implementation overhead; (3) we demonstrate that our methods accelerate optimization on tasks such as hyperparameter search. 

\section{BACKGROUND}

\subsection{Calibrated \& Conformal Prediction}
Traditionally, predictive uncertainty in statistics is evaluated using proper scoring rules \citep{gneiting2007strictly}. Proper scoring rules measure precisely two characteristics of a forecast: calibration and sharpness \citep{murphy1987general}.
Intuitively, calibration means that a 90\% confidence interval contains the outcome  about $90\%$ of the time.
Sharpness means that confidence intervals should be tight. Maximally tight and calibrated confidence intervals are Bayes optimal.

\paragraph{Calibration.}
Formally, suppose we have a model $M:\mathcal{X} \to \mathcal{P}(\mathbb{R})$ that outputs a probabilistic forecast $Q_x \in \mathcal{P}(\mathbb{R})$ of a target variable $y \in \mathbb{R}$ given an input $x \in \mathcal{X}$. 
In this paper, we will assume that $Q_x$ is represented by a quantile function.
%
When the target $y \in \mathbb{R}$ is continuous, calibration is often defined as
$P(Y \leq Q_{X}(p)) = p , \; \forall p \in [0,1]$ \citep{kuleshov2018accurate}.
In an online setting, this definition becomes
\begin{equation}
\frac{\sum_{t=1}^T \mathbb{I}\{ y_t \leq Q_t(p) \}}{T} \to p \; \textrm{for all $p \in [0,1]$} \label{eqn:calibration1}
\end{equation}
as $T \to \infty$, where $\mathbb{I}$ is the indicator function and $Q_t = M(x_t)$ is the forecast at time $t$. 

\paragraph{Calibrated Prediction} Out of the box, most models $M$ are not calibrated. Post-hoc calibration and conformal prediction yield calibrated forecasts by adjusting predicted probabilities $Q$ from $M$ on a held out dataset \citep{tutorialconformal2007shafer,kuleshov2018accurate,VovkPTGAC20}. 

For training a recalibrator over our probabilistic model, we compute the CDF $F_t$ at each data-point $y_t$ using the formulation $F_t=[\mathcal{M}(x_t)](y_t)$. This can be used to estimate the the empirical fraction of data-points below each quantile. Algorithm~\ref{alg:train-recalib-general} based on based on \citet{kuleshov2018accurate} outlines this procedure.
\begin{algorithm}
  \caption{Calibration of Probabilistic Model 
  }
  \label{alg:train-recalib-general}
  \textbf{Input:} Dataset of probabilistic forecasts and outcomes $\{[\mathcal{M}(x_t)](y_t), y_t\}_{t=1}^{N}$ 
  \begin{enumerate}
      \item Form recalibration set 
      $\mathcal{D} = \{[F_t, \hat{P}(F_t)\}_{t=1}^{N}$
      where $F_t=[\mathcal{M}(x_t)](y_t)$ and $\hat{P}(p) = |\{ y_t | [F_t\leq p, t=1,..,N\}|/N$.
      \item Train recalibrator model $\mathcal{R}$ on dataset $\mathcal{D}$.
  \end{enumerate}
\end{algorithm}

\subsection{Bayesian Optimization}

Uncertainty estimation plays an important role in sequential decision-making, where we observe a sequence of inputs $x_t \in \mathcal{X}$ for $t=1,2,...,T$ and choose actions $a_t \in \mathcal{A}$, after which nature reveals outcomes $y_t$. In such settings, an accurate probabilistic model of $y_t$ given $x_t$ is useful for choosing $a_t$.

Bayesian optimization is a sequential decision-making process that seeks to find a global minimum $x^\star \in \arg \min_{x \in \mathcal{X}} f(x)$ of an unknown black-box objective function $f:\mathcal{X} \to \mathbb{R} $ over an input space $\mathcal{X} \subseteq \mathbb{R}^D$. Please note that $f$ can be deterministic or non-deterministic. 
%
Computing $f(x)$ is usually computationally expensive; furthermore we may not have access to the values of $f$ or its gradient. A classical application area of Bayesian optimization is hyperparameter search, where $x\in \mathcal{X}$ are choices of hyperparameters, and $f(x)$ is the resulting performance of a machine learning model. 

At each step t, Bayesian optimization forms a probabilistic model $M_t:\mathcal{X} \to \mathcal{P}(\mathbb{R})$ of $f$; the output of $M_t(x)$ is a probability distribution over the value of $f(x)$. 
%
We use uncertainty estimates from this probabilistic model to pick $x_t$ and we update $M_t$. 
%


 \begin{algorithm}[h]
\SetAlgoLined
 Initialize model $M_0$ with data $\mathcal{D}_0=\{x_t, y_t\}_{t=1}^{N}$\;
 \For{$t=1,2,...,T$}{
  $x_t$ = $\arg \max_{x \in \mathcal{X}}\textrm{Acquisition}(x, M_{t-1});$\\
  $y_t$ = $f(x_t);$\\
  $\mathcal{D}_t$ = $\mathcal{D}_{t-1} \cup \{(x_t,y_t)\}$;\\
  Construct model $M_t$ on data $\mathcal{D}$;
 }
 \caption{Bayesian Optimization}
 \label{algo-bo}
\end{algorithm}

Above, $\mathrm{Acquisition}(x, M)$ is an acquisition function; common examples include expected improvement, probability of improvement, and upper counfidence bounds (UCB)~\citep{frazier2018tutorial}.



\section{UNCERTAINTY IN BAYESIAN OPTIMIZATION} 


\subsection{Which Uncertainties Are Needed in Online Decision-Making?}

Online sequential decision-making tasks such as Bayesian optimization benefit from an accurate probabilistic model to determine which actions to choose. However, because data is limited and because of the need to make modeling assumptions, most predictive models are not optimal. This raises the question: which aspects of a predictive model are important for a downstream decision-making task?



This paper argues that the calibration-sharpness tradeoff has important implications on downstream performance. In particular, we argue that among models that attain a given proper loss, it is {better to achieve good levels of calibration}. 



\paragraph{Why is Calibration Useful?}

A key challenge faced by decision-making 
algorithms is balancing exploration---e.g., learning the shape of the unknown function $f$ in Bayesian optimization---against exploitation---e.g., selecting points $x$ at which $f$ takes small values.
%
Exploration-exploitation decisions are often made using a probabilistic model. In regions that are unexplored, the confidence interval around the value of $f(x)$ should be large to 
promote exploration.
%
Calibration helps mitigate over-confidence and promotes accurate confidence intervals that encourage exploration.

Another benefit of calibrated models is the accurate computation of expected values of future outcomes. Since an expectation is a sum weighted by probabilities of future events, aligning predicted and empirical probabilities is crucial. Accurate estimates of expected utility yield improved planning performance in model-based algorithms \citep{malik2019calibrated}.


\subsection{Formal Analysis}

\textbf{Notation.} Consider a setting where we sequentially minimize a loss function $\ell : \mathcal{Y} \times \mathcal{A} \times \mathcal{X} \to \mathbb{R}_+$ over a set of outcomes $\mathcal{Y}$, actions $\mathcal{A}$, and features $\mathcal{X}$. The loss $\ell(y,a,x)$ quantifies the error of an action $a \in \mathcal{A}$ in a state $x \in \mathcal{X}$ given outcome $y \in \mathcal{Y}$.

Bayesian decision-making theory provides a principled approach for selecting actions in the above scenario. We rely on a predictive model $M$ of $y$ and select decisions that minimize the expected loss:
\begin{align}
a(x) & = \arg \min_a \mathbb{E}_{y \sim M(x)} [ \ell(y, a,x) ] \\
\ell(x) & = \min_a \mathbb{E}_{y \sim M(x)} [ \ell(y, a,x) ].
\end{align}
Here, $a(x)$ is the action that minimizes the expected loss under $M$.
%
%
If $M$ were a perfect predictive model, the above decision-making strategy would be optimal. In practice, inaccurate models can yield imperfect decisions. 
We argue below that in some cases, calibration is a weaker condition that helps accurately estimate the value of a loss function.
%

Specifically, we provide a concentration inequality on estimates of the loss that generalizes results by \citet{zhao2020individual} in the i.i.d.~setting.
Our result holds for losses $\ell(y,a,x)$ that are monotonically non-increasing or non-decreasing in $y$. 
Note that common acquisition functions used in Bayesian optimization yield $\ell$ that satisfy this condition.

\begin{restatable}{theorem}{markovvariant}
\label{thm:dist_calib_bound}
Let $M$ be a quantile calibrated model as in (\ref{eqn:calibration1}) and
let $\ell(y, a, x)$ be a monotonic loss.
Then for any sequence $(x_t, y_t)_{t=1}^T$ and $r > 1$, we have:
\begin{equation}
    \label{eqn:dist_calib_bound1}
    \lim_{T \to \infty} \frac{1}{T} \sum_{t=1}^T \mathbb{I} \left[ \ell( y_t, a(x_t), x_t) \geq r \ell(x_t)) \right] \leq 1 / r
\end{equation}
\end{restatable}
Note that this statement represents an extension of Markov's inequality. 
See Appendix~\ref{apdx:math_proofs} for a proof.

\section{ALGORITHMS FOR ONLINE CALIBRATION}

Enforcing calibration in an online decision-making setting is challenging  because the data distribution is non-stationary (it is influenced by the agent's decisions).
Note that the objective itself is stationary and fixed; however, the distribution of the data at each timestep depends on the output of the algorithm at previous timesteps, hence is non-IID (see Appendix~\ref{apdx:stationarity-clarification}). 
We address the issue of non-stationary data distribution by introducing new algorithms based on online learning \citep{shalev2012online}.
Online learning provides methods that provably produce accurate predictors on any stream of datapoints, including data chosen by an adversary.

\subsection{Online Recalibration}

\paragraph{Setup}
At each time step $t=1,2,..., T$ we observe features $x_t \in \mathcal{X}$. A predictive model produces a forecast $Q_t : [0,1] \to \mathbb{R}$ based on $x_t$ that takes the form of a quantile function.
We assume that $Q_t$ is invertible and use the convention that $Q_t(p) = \infty$ for $p > 1$ and $Q_t(p) = -\infty$ for $p < 0$. As an example, a Gaussian process model \citep{Rasmussen2005GPML} used for Bayesian optimization outputs forecasts $Q_t$ given by the quantile function of a Gaussian distribution.

Initially, the forecasts $Q_t$ may be miscalibrated; we seek to compose $Q_t$ with a recalibrator $R_t : [0,1] \to [0,1]$ such that $Q_t \circ R_t$ is calibrated as in (\ref{eqn:calibration1}). After we choose $Q_t \circ R_t$, nature reveals a label $y_t \in \mathbb{R}$; we use $o_t(y_t, p) = \mathbb{I}\{ y_t \leq Q_t(p) \}$ to denote the indicator of the outcome that $y_t$ falls below the $p$-th quantile. Our goal can be equivalently defined as choosing $R_t$ such that,
\begin{equation}
    \frac{1}{T}\sum_{t=1}^T o_t(y_t, R_t(p)) \to p \text{ as $T \to \infty$ $\forall p \in [0,1]$}.
    \label{eqn:cal}
\end{equation}
\paragraph{Algorithms}

Our strategy will be to construct $R_t$ by {\em optimizing for calibration on historical data}.
Specifically, we choose $R_t$ such that 
\begin{align}
R_t(p) \in \arg\min_q \left[ \psi(q) + \sum_{s=1}^{t-1} \ell_{sp}(y_s, q) \right],
\label{eqn:ftrl}
\end{align}
where $\ell_{sp} : \mathbb{R} \times [0,1] \to \mathbb{R}_+$ is a loss function 
(possibly varying in time $s$) that quantifies miscalibration and $\psi(y) : \mathbb{R} \to \mathbb{R}_{+}$ is a regularizer. 
Note that the loss function $\ell_{sp}$ is internal to the recalibrator.  We refer to the dataset used to define $R_t$ (i.e., the dataset over which we compute the $\arg\min$) as the {\bf calibration set}; it consists of forecasts $Q_s$ and outcomes $y_s$, and is denoted as%
\begin{equation}
    \mathcal{C}_t = \{(Q_s, y_s) \mid s = 1,2,..., t-1\}.
    \label{eqn:create-cal-set}
\end{equation}
Normally $\mathcal{C}_t$ consists of data from all timesteps $s < t$; we will discuss additional ways of constructing $\mathcal{C}_t$ below.

Equation \ref{eqn:ftrl} establishes a close connection to online optimization: it implements a classical online learning algorithm called {\em follow the regularized leader} (FTLR) \citep{shalev2012online}.
Below, we will adopt $\ell_{sp}$ that are derived from the pinball loss---a generalization of the L1 loss motivated by conditional quantile estimation. 

\subsection{Recalibration via Online Optimization}



Consider first the simpler problem of finding a $q_t \in [0,1]$ such that $Q_t(q_t)$ is an estimate of the $p$-th conditional quantile, i.e., 
$
\frac{1}{T}\sum_{t=1}^T o_t(y_t, q_t) - p \to 0 \text{ as $T \to \infty$}.
$

\paragraph{Quantile Pinball Loss}
We propose to optimize a modification of the pinball loss which we call the quantile pinball loss (QPL), defined as
\begin{align}
    \ell_{tp}(y_t, q) 
    & = (q-Q_t^{-1}(y_t))(o_t(y_t, q) - p).
    \label{eqn:quantile_pinball}
\end{align}
Note that $\ell_{tp}$ is convex: its graph is V-shaped with the slopes of the two lines defining the V being $p$ and $1-p$; when $p=0.5$, this essentially yields the L1 loss.   
The offline minimizer of the pinball loss yields a consistent estimator for the $p$-th quantile of the data in a batch setting \citep{koenker1978regression}; here, we derive the same property in the online setting. 

\begin{lemma} 
    The quantile pinball loss serves as a quantile estimator, in that $\arg\min_q  \sum_{s=1}^t \ell_{sp}(y_s, q)$ over a dataset $(y_s)_{s=1}^T$ yields a $p$-th quantile of the dataset.
\end{lemma}
Please refer to Appendix~\ref{sec:app_algorithms} for a complete proof and discussion of QPL. 

\paragraph{Optimization}
Our proposed algorithm optimizes the QPL using a classical online learning algorithm called online gradient descent (OGD). This algorithm may be defined as optimizing (\ref{eqn:ftrl}) for a specific choice of loss function (thus it is an instance of the follow-the-regularized leader framework). Specifically, we define $q_t$ at each step as
\begin{equation}
    q_t \in \arg\min_q \left[ \frac{1}{2\eta} q^2 + \sum_{s=1}^{t-1} \overline{\ell}_{sp}(q) \right], \label{eqn:ftrl2}
\end{equation}
where $\overline{\ell}_{tp}(q)$ is the linearization of the QPL at $q=q_t$, defined as \begin{align*}
    \overline{\ell}_{tp}(q) = (q - q_t) (o_t(y_t, q_t) - p).
\end{align*}
Here, we have dropped an additive constant from the linearization since this term does not involve $q$ and therefore has no effect on the minimization problem in Equation \eqref{eqn:ftrl2}. 
Computing $q_t$ can be as simple as a convex optimization problem over a scalar in $[0,1]$ (solvable using e.g., binary search). 

The linearization $\overline{\ell}_{tp}(q)$ approximates $\ell_{tp}(y_t, q)$ everywhere by the supporting hyperplane given by a subgradient at $q=q_t$. 
Since $\psi(q) = \frac{1}{2\eta} q^2$, we can also set the derivative of the objective in (\ref{eqn:ftrl2}) to zero and show that $q_t = \sum_{s=1}^{t-1} \eta g_s$ for $g_s \in \partial_q \ell_{sp}(q_s)$, where $\partial_q \ell_{sp}(q_s)$ is the subgradient at the points $q_s$ for $s < t$. 
This derivation shows that our proposed algorithm is equivalent to online subgradient descent (OSD) on $q$
\citep{hazan2016introduction}, and reveals the ties between our approach and online learning.

\paragraph{Quantile Calibration}

Online learning algorithms such as OSD and FTRL guarantee vanishing regret on any sequence of $(x_t, y_t)$. Here, we also show that these algorithms yield quantile calibration. Interestingly, our proof does not directly rely on regret minimization.

\begin{restatable}{theorem}{consistentsingle}
\label{thm:consistentsingle}
    For any sequence $(Q_t, y_t, q_t)_{t=1}^T$, where $q_t$ satisfies \eqref{eqn:ftrl2} with $\eta > 0$ and $p \in [0,1]$, we have
    \begin{equation}
        \left| \frac{1}{T}\sum_{t=1}^T o_t(y_t, q_t) - p  \right|\leq \frac{1 + \eta }{\eta T}.
    \end{equation} \label{thm:one_quantile_calibration}
\end{restatable}
\begin{proof}[Proof (Sketch)]
    We can show that for any $T$, $-\eta \leq q_T \leq 1+\eta$.
    Observe also from (\ref{eqn:quantile_pinball}) that the subgradient $g_t \in \partial \ell_t(q_t)$ can be written as $(o_t(y_t, q_t)) - p)$. Since
    $q_{T+1} = \sum_{t=1}^{T-1} \eta g_t$, and $q_{T+1} \in [-\eta, 1+\eta]$, we get the desired result by substituting in the previous expression for $g_t$ and dividing both sides by $\eta T$.
\end{proof}

See Appendix \ref{sec:app_algorithms} for the full proof.
Our result is reminiscent of adaptive conformal inference \citep{gibbs2021adaptive}. However, key differences include: (1) our approach draws novel connections to online learning; 
(2) our results naturally generalize to full quantile recalibration; (3) our results hold as $\eta \to \infty$. 

\paragraph{Quantile Function Recalibration}

We may now define a full recalibrator $R_t$ pointwise as
\begin{equation}
    R_t(p) = \inf \arg\min_q \left[ \frac{1}{2\eta} q^2 + \sum_{s=1}^{t-1} \bar\ell_{sp}(y_t, q) \right], \label{eqn:full_r}
\end{equation}
where $\bar\ell_{sp}$ is defined as in (\ref{eqn:ftrl2}); the $\inf$ is used to break ties if the $\arg\min$ is a set. 
It is easy to compute $R_t(p)$ at any $t, p$ by solving (\ref{eqn:full_r}), and we can also approximate $R_t$ via a linear interpolation at a number quantiles. The correctness result below follows from Theorem \ref{thm:one_quantile_calibration} and the properties of the quantile pinball loss; see Appendix \ref{sec:app_algorithms} for the proof.

\begin{restatable}{theorem}{consistencyall}
\label{thm:consistencyall}
    For any sequence $(Q_t, y_t)_{t=1}^T$, learning rate $\eta > 0$, 
    and quantile $p \in [0,1]$,
    $$
    \left| \frac{1}{T}\sum_{t=1}^T\mathbb{I}\{ y_t \leq Q_t(R_t(p)) \} - p  \right|\leq \frac{1 + \eta }{\eta T}.
    $$
\end{restatable}

%

\section{CALIBRATED BAYESIAN OPTIMIZATION}

Next, we apply our online calibration algorithms to a sequential decision-making task: Bayesian optimization.
%
Algorithm~\ref{calibrated-bo} outlines our proposed procedure.
At each step $t$, we compose the Bayesian optimization model $M_t$ (which we can assume without loss of generality as being represented by a quantile function) with a recalibrator $R_t$.
We define $R_t$ as in (\ref{eqn:full_r}); we denote the resulting subroutine as $\textsc{Calibrate}(M_t, \mathcal{D}_t)$.

\begin{algorithm}[h]
\SetAlgoLined
 Initialize model $M_0$ with data $\mathcal{D}_0=\{x_t, y_t\}_{t=1}^{N}$\;
 $R_0 = \textsc{Calibrate}(M_0, \mathcal{D}_0)$\;
 \For{$t=1,2,...,T$}{
  $x_t$ = $\arg \max_{x \in \mathcal{X}}\textrm{Acquisition}(x, M_{t-1} \circ R_{t-1})$\;
  $y_t$ = $f(x_t)$\;
  $\mathcal{D}_t$ = $\mathcal{D}_{t-1} \cup \{(x_t,y_t)\}$\;
  Construct model $M_t$ on data $\mathcal{D}$\;
  $R_t = \textsc{Calibrate}(M_t,\mathcal{D}_t)$\;
 }
 \caption{Calibrated Bayesian Optimization}
 \label{calibrated-bo}
\end{algorithm}



\paragraph{Practical Considerations.}

We also propose an additional heuristic for constructing the calibration set $\mathcal{C}_t$ (Algorithm~\ref{alg:recal}). Our heuristic is motivated by the fact that Bayesian optimization is typically run for a short number of steps $T$ and offers further performance gains (see Section \ref{sec:benchmarks}).


\begin{algorithm}[h]
  \caption{{$\textsc{Calibrate}$}}
  \label{alg:recal}
  \textbf{Input:}  Model $M$, Dataset $\mathcal{D}=\{x_t, y_t\}_{t=0}^{N}$ \;
    Initialize recalibration dataset $\mathcal{C} = \emptyset$ \;
    $S = 
    \textsc{CreateSplits({D})}$ \;
    For each $(\mathcal{D_{\textrm{train}}}, \mathcal{D_{\textrm{test}}})$ in $S$: \\
          \hspace{2mm} 1. Train base model $M$ on dataset $\mathcal{D_{\textrm{train}}}$ \;
          \hspace{2mm} 2. Create calibration set $\mathcal{C}'$ from $\mathcal{D_{\textrm{test}}}$ \;
          \hspace{2mm} 3. $\mathcal{C} = \mathcal{C} \cup \mathcal{C}'$ \;
      Return recalibrator $R$ obtained via (\ref{eqn:full_r}) on $\mathcal{C}$\;
\end{algorithm}

Specifically, Algorithm~\ref{alg:recal} builds a recalibration dataset $\mathcal{D}_\text{recal}$ via cross-validation on $\mathcal{D}$. 
At each step of cross-validation, we train $M$ on the training folds and compute forecasts on the test folds, thus generating a calibration dataset (\ref{eqn:create-cal-set}) using $\mathcal{D_{\textrm{test}}}$. The union of all the forecasts on the test folds produces the final calibration dataset on which the recalibrator is trained. 
In our experiments, we used leave-one-out cross-validation within $\textsc{CreateSplits}$.
Formally, given a dataset $\mathcal{D} = \{x_t, y_t\}_{t=1}^N$, $\textsc{CreateSplits}(\mathcal{D})$ produces $N$ splits of the form $\{(\mathcal{D} \; \backslash \; \{x_i, y_i\}, \{x_i, y_i\}) \mid i=1,2,...,N\}$.
%



\paragraph{Understanding Acquisition Functions}

Next, we discuss how calibration is useful in combination with various acquisition functions.
The {\bf probability of improvement} can be written as ${1 - F(f(x_\text{new})+\epsilon)}$, where $F$ is the CDF predicted by the model. In a calibrated model, this predicted probability in the long run matches the empirical probability of observing an improvement.
The {\bf expected improvement} can be defined as ${\mathbb{E}[\max(f(x_\text{old}) - f(x_\text{new}), 0)]}$. 
Theorem \ref{thm:dist_calib_bound} suggests that this expectation is easier to evaluate under a calibrated model.
{\bf Upper confidence bounds} are ${\mu(x) + \alpha\cdot \sigma(x)}$ for Gaussians and $Q_x(\alpha)$ for general $Q$, i.e., the $\alpha$-th quantile. Recalibration ensures that $Q_x(\alpha)$ is truly the $\alpha$-th quantile (it is above the true $y$ a fraction $\alpha$ of the time). Appendix~\ref{apdx:acquisition_functions} discusses acquisition functions further. 


\section{EXPERIMENTS}

We perform experiments on several benchmark objective functions that are standard in the Bayesian optimization literature, as well as on a number of hyperparameter optimization tasks.
\begin{figure}[tb]
\subfigure[Uncalibrated]{
  \includegraphics[scale=0.24, trim=0 0 0 1.3cm,clip]{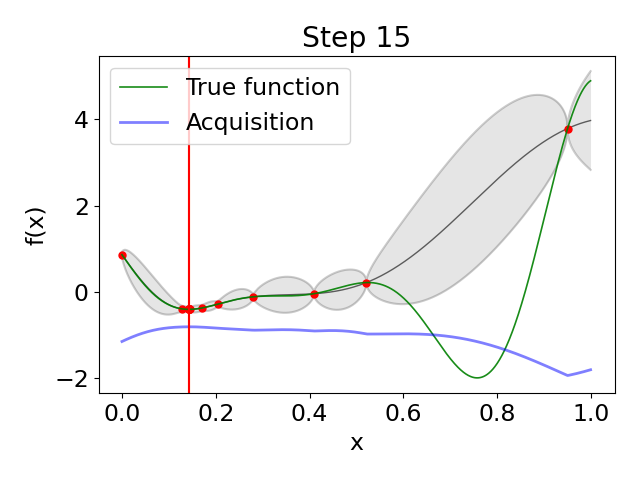}}
  \subfigure[Calibrated]{
  \includegraphics[scale=0.24, trim=0 0 0 1.3cm,clip]{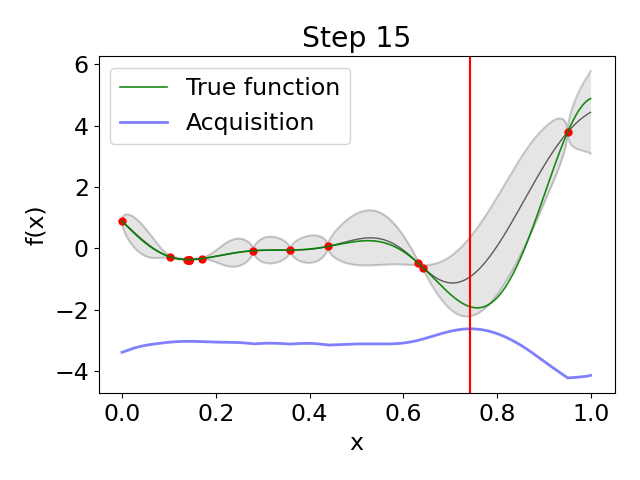}}
  \caption{Comparison of Uncalibrated and Calibrated Bayesian Optimization on the Forrester Function (Green) Using the UCB Acquisition Function (Blue).}
  \label{fig:visualize-forrestor}
\end{figure}
\begin{figure*}[!htb]
\centering     
\subfigure[Forrester]{\label{fig:forrestor}\includegraphics[width=0.3\linewidth]{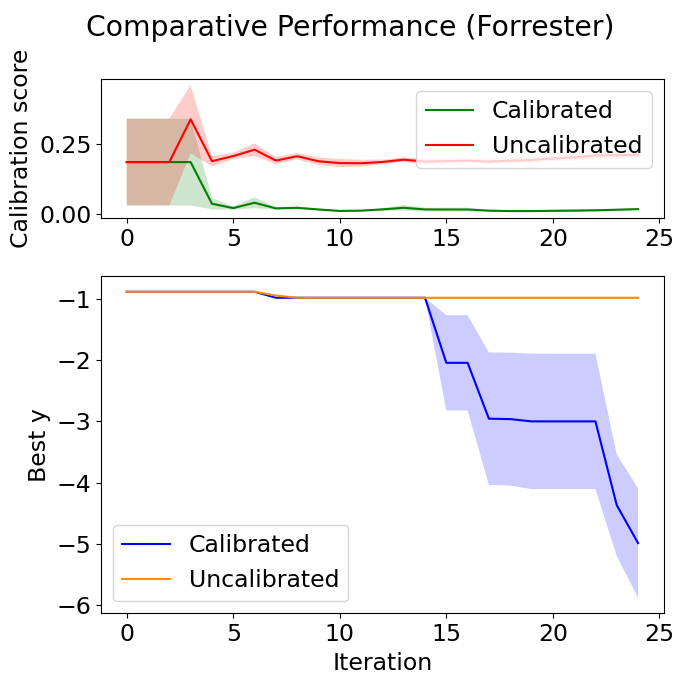}}
\subfigure[Ackley (2D)]{\label{fig:ackley}\includegraphics[width=0.3\linewidth]{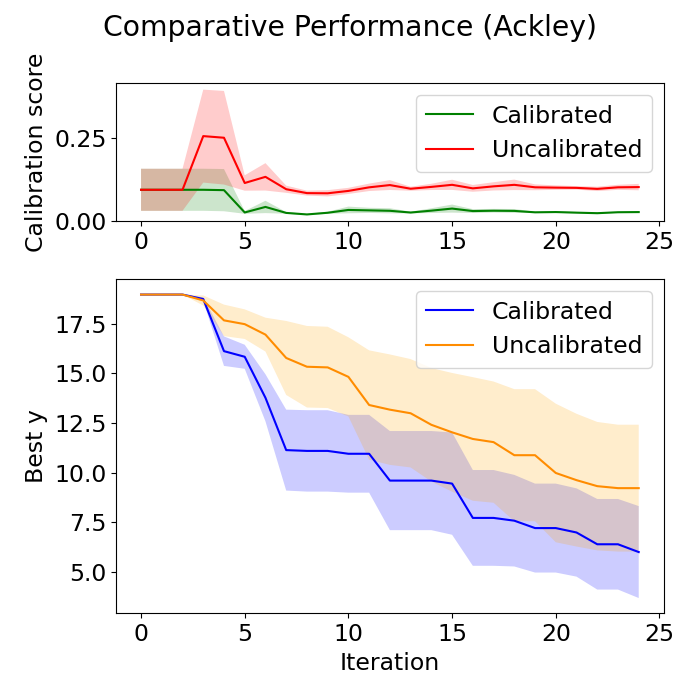}}
\subfigure[Alpine (10D)]{\label{fig:alpine1_10d}\includegraphics[width=0.3\linewidth]{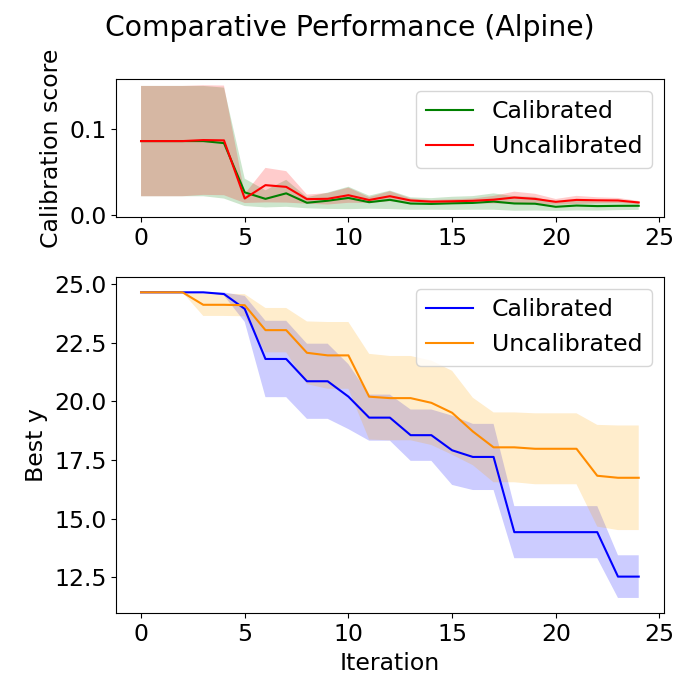}}
\caption{Comparison of Bayesian Optimization in Benchmark Functions. In the top plots, we see that calibrated method reduces calibration error. The bottom plots show that on an average, the calibrated method identifies the minimum using less iterations.
}
\label{benchmark_comparison}
\end{figure*}
\paragraph{Setup.}
We use the Gaussian Process (GP) as our base model in the following experiments. However, our method can be applied to any probabilistic model underlying Bayesian optimization in general \citep{pmlr-v37-snoek15},  ~\citep{NIPS2016_a96d3afe}. 
See Appendix~\ref{apdx:experimental_details} for additional implementation details.
 \textbf{Analysis of Calibration.} We assess the calibration of the original probabilistic model underlying Bayesian optimization using calibration scores as defined by ~\cite{kuleshov2018accurate}. Thus,
$\text{cal}(F_1, y_1,..,F_n, y_n) = \sum_{j=1}^{m} (p_j - \hat{p_j})^2,$
where $0\leq p_1 < p_2 <..<p_m \leq 1$ are $m$ confidence levels we use to compute the calibration score. $\hat{p_j}$ is estimated as $\hat{p_j} = |\{ y_t | [F_t\leq p_j, t=1,..,N\}|/N.$ The calibration scores are computed on a test dataset $\{F_t, y_t\}_{t=0}^{T}$. This test dataset is constructed by setting $F_t=F_{x_\textrm{next}}(y_\textrm{next})$ and $y_t=y_\textrm{next}$ at every step $t$ before updating model $\mathcal{M}$ in Algorithm~\ref{calibrated-bo}. In our experiments, we average the calibration score at every step of Bayesian optimization over 5 repetitions of the experiment for each benchmark function.
 
\subsection{Benchmark Optimization Tasks}
\label{sec:benchmarks}

We visualize runs of calibrated and plain Bayesian optimization on a simple 1D task --- the Forrester function in Figure~\ref{fig:visualize-forrestor}. 
We use the Upper Confidence Bound (UCB) as our acquisition function. 
Both functions start at the same three points, which miss the global optimum in $[0.6, 0.9]$. 
The base GP is overconfident, gets stuck at a local minimum near 0.2, and never explores the optimal region. 
However, the calibrated method learns that its confidence intervals are too narrow and expands them. 
This leads it to quickly identify the global optimum in $[0.6, 0.9]$. 
Please refer to Appendix~\ref{apdx:visualize_forrester} for additional plots. 

\begin{figure*}[h]
\centering     
\subfigure[UCB]{\label{fig:sixhumpcamel_lcb}\includegraphics[width=0.3\linewidth]{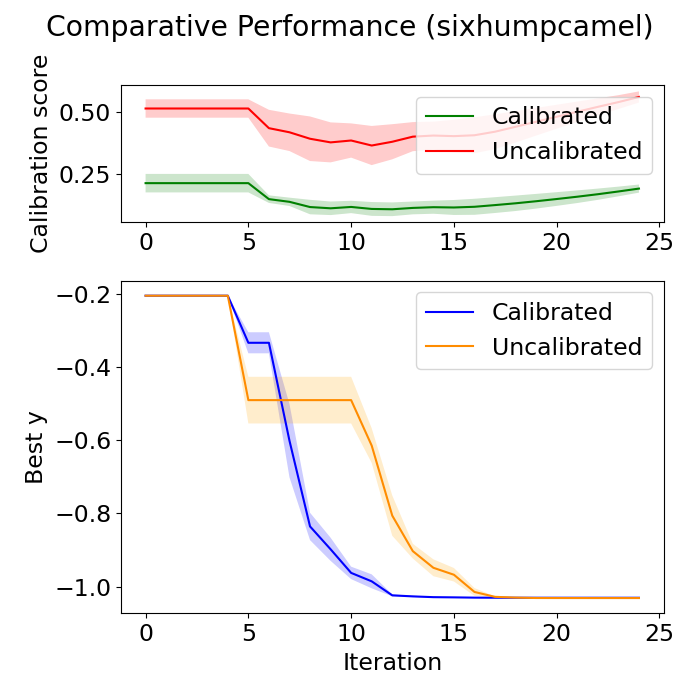}}
\subfigure[EI]{\label{fig:sixhumpcamel_ei}\includegraphics[width=0.3\linewidth]{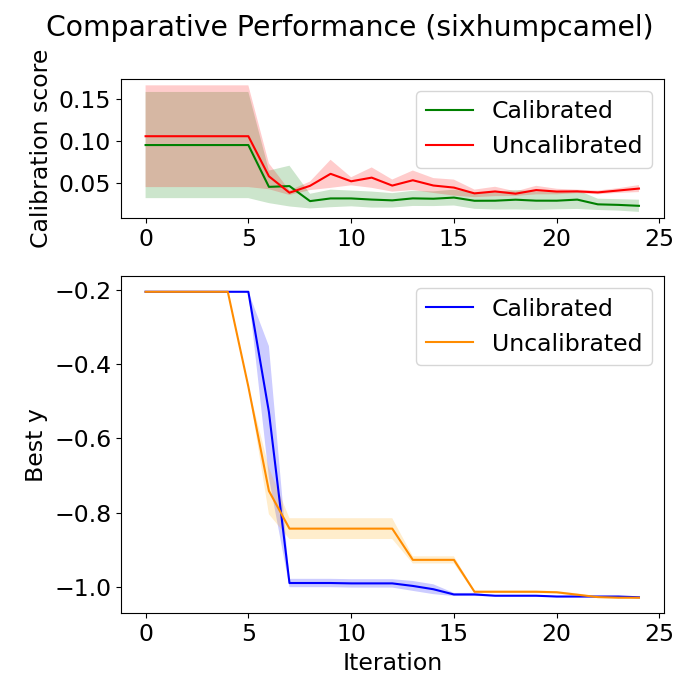}}
\subfigure[PI]{\label{fig:sixhumpcamel_pi}\includegraphics[width=0.3\linewidth]{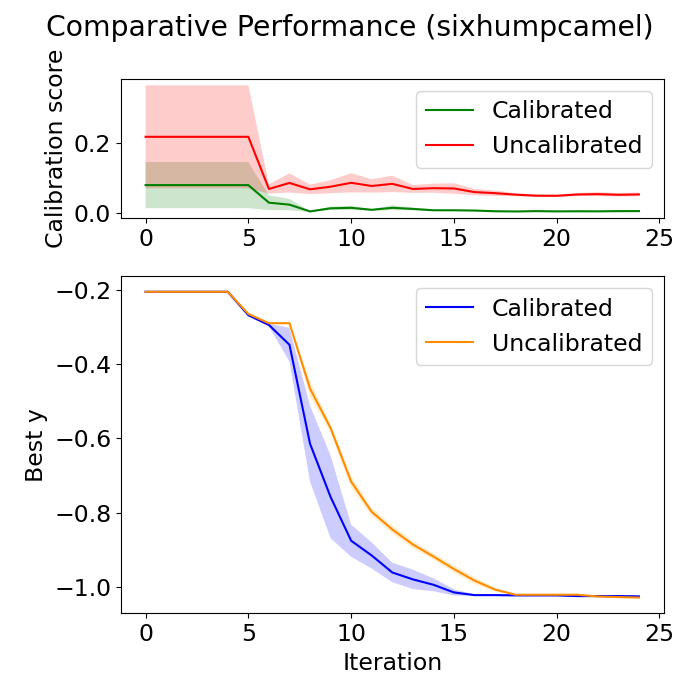}}
\caption{
Comparison of Calibrated and Uncalibrated Bayesian Optimization on Six-hump-camel Function (2D) for Various Acquisition Functions. The top plots show that the calibrated method reduces calibration error. In the bottom plots, we see that the calibrated method identifies the minimum using fewer iterations.
}
\label{fig:sixhumpcamel_final}
\end{figure*}

In Figure~\ref{fig:forrestor}, we compare calibrated and uncalibrated Bayesian optimization on the Forrester function. 
In Figure~\ref{fig:ackley} and Figure~\ref{fig:alpine1_10d}, we compare the performance of calibrated method against uncalibrated method under EI acquisition function on the 2D Ackley function and 10D Alpine function \citep{simulationlib}. In Appendix~\ref{apdx:increasing-optimization-step-num}, we also run the 10D Alpine function for a greater number of steps (100) and continue to see that the calibrated method produces a lower minima. 
In Figure~\ref{fig:sixhumpcamel_final}, we compare the performance of our method on the Sixhump Camel function while varying the acquisition function.
In all these examples, the calibrated method finds the global minimum before the uncalibrated method on average. 

\begin{figure*}[h]
\vspace{-0.3cm}
\centering
  \includegraphics[scale=0.24]{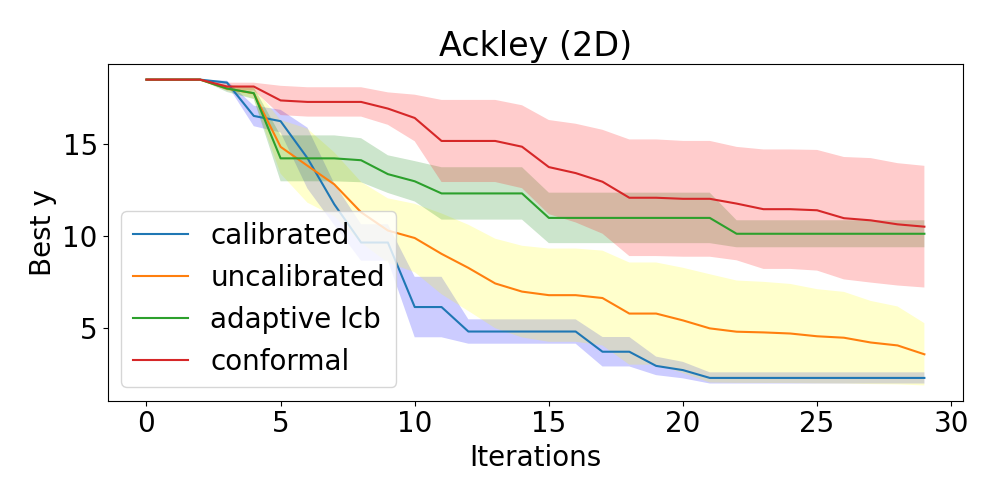}
  \includegraphics[scale=0.24]{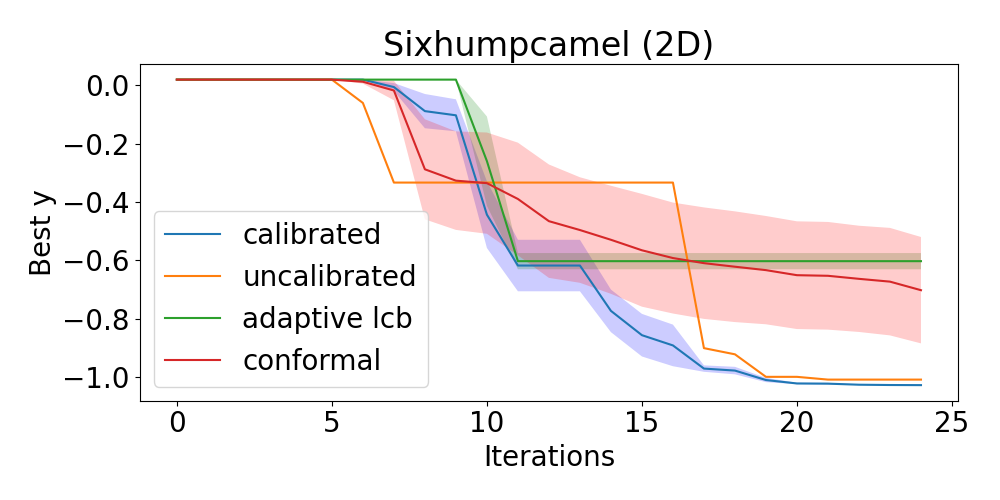}
\caption{Comparison of Calibrated Bayesian Optimization with Additional Baselines. The calibrated method outperforms the modern Bayesian optimization baseline~\citep{stanton2023bayesian} and Adaptive UCB for Bayesian optimization with unknown kernel hyperparameters~\citep{berkenkamp2019noregret}. The acquisition function used here is UCB.}
  \label{fig:conformal-baselines}
  \vspace{-0.3cm}
\end{figure*}

\textbf{Comparison Against Additional Baselines.}
In Figure~\ref{fig:conformal-baselines}, we see that our calibrated method compares favorably against the modern conformal Bayesian optimization~\citep{stanton2023bayesian} and Adaptive UCB for Bayesian optimization with unknown kernel hyperparameters~\citep{berkenkamp2019noregret}. Please refer to Appendix~\ref{apdx:baseline-comparison} for additional results.

In Table~\ref{table:baselines}, we compare our method against baselines including input and output warping \citep{snoek2014input, NIPS2003_6b5754d7}, Box-Cox and log-transforms on the output \citep{rios2018} and an ensemble of Gaussian processes \citep{lakshmi2016simple}. 
Metrics used to perform this evaluation include the minimum value $m$ of the objective function achieved by Bayesian optimization, fraction $f$ of experimental repetitions where the baseline performs worse than our method and normalized area under the Bayesian optimization curve \textbf{$a$}. We show the error bars in braces for the minimum value $m$. Although $f$ is not amenable to the same error bar computation, we can estimate the variance using analytical formula for Bernoulli($p$) as $p(1-p)$. Also, the error bars on reported AUC values are $<0.02$. 
Please refer to Appendix~\ref{apdx:evaluation_metrics} for detailed definition of these metrics.  

In Table~\ref{table:baselines}, we see that warping methods occasionally improve over the uncalibrated method, often worsen it, and recalibration is almost always more effective than warping. Both uncertainty estimation methods (ensembling and calibration) improve performance relative to the uncalibrated method, with calibration yielding the bigger improvement.

\begin{table*}[htb]
  \caption{Comparison of Calibrated Bayesian Optimization Against Baselines.}
  \label{table:baselines}
  \centering
  \resizebox{\textwidth}{!}{
  
  \begin{tabular}{lllcc}
    
    Objective & Method	& Minimum Found & Fraction of Runs Where  & 	Area Under the Curve \\
    	Function & & (↓) & Calibrated Beats Baseline (↑) & 	 (↓)\\
    \midrule
    
    Forrester (1D) 
    &Uncalibrated method  & -0.986 (0.001) &	0.8 & 0.9866\\
    &Input warping & -0.889 (0.000) &\textbf{1}& 1.0000\\
    &Output warping & \textbf{-6.021 (0.000)} & 0.0 & \textbf{0.2409}\\
    &Boxcox transformation of output & -4.806 (0.780) & 0.4 & 0.4911\\
    &Log transformation of output & -0.986 (0.000) & \textbf{1} & 0.9865\\
    &Bagged ensemble of 5 GPs & -3.891(1.098) & 0.8& 0.8382\\
    &Calibrated Method (Ours) & -4.983 (0.894)&	\textbf{1}	& 0.8187\\
    \midrule

    Ackley (2D) 
    &Uncalibrated method    & 12.359 (2.439) & 0.8	& 0.7815\\
    &Input warping  & 14.061 (1.312) & 0.8	& 0.8125\\
    &Output warping      & 10.459 (3.365) & 0.6 &	0.7257     \\
    &Boxcox transformation of output   & 14.421 (1.423)&	0.8	&0.9015     \\
    &Log transformation of output & 8.330 (0.849)&	0.8	& 0.6524     \\
    &Bagged ensemble of 5 GPs    & 6.011 (2.544)	& 0.6	& 0.6013   \\
    &Calibrated Method (Ours) & \textbf{5.998 (2.314)} & \textbf{1} &	\textbf{0.5516}\\
    \midrule
        
    Alpine (10D)
    &Uncalibrated method  &15.506 (1.275) &	0.6 &	0.6527\\
    &Input warping  & 15.697 (1.740) &	0.8 &	0.6492\\
    &Output warping  & 13.531 (2.127)	& 0.6 &	\textbf{0.5857}\\
    &Boxcox transformation of output  & 15.715 (0.603)	& 0.8 &	0.6253\\
    &Log transformation of output  & 20.996 (1.661)	& 0.8 &	0.7931\\
    &Bagged ensemble of 5 GPs  & 16.677 (1.699)	& 0.8 &	0.7334\\
    &Calibrated Method (Ours) & \textbf{12.537 (0.909)}	& \textbf{1}	& 0.6423\\
    \bottomrule
\end{tabular}
  }
\end{table*}

\begin{table*}[htb]
  \caption{Evaluating Calibrated Bayesian Optimization on Additional Tasks. Calibration strictly improves performance on four benchmarks and performance is similar on two tasks. Lower performance on Cross-in-Tray could be attributed to the presence of multiple sharp edges and corners that are hard to model via a GP.}
  \label{table:additional-benchmarks}
  \centering
  \begin{small}
  {
  \begin{tabular}{lccc}
    Optimization & \% Runs Where &	Area Under the Curve, 	&Area Under the Curve, \\
    benchmark	& Calibrated is Best (↑)&	Calibrated (↓)	& Uncalibrated (↓)\\
    \midrule
    Cosines	& 0.8	& 0.2973	& 0.3395\\
    Beale	& 0.6 &	0.0929 &	0.0930\\
    Mccormick	& 0.8 & 0.1335 & 0.1297\\
    Powers	& 0.8 & 0.2083 & 0.2325\\
    Cross-in-Tray & 0.2 & 0.2494 & 0.2217\\
    Ackley	& 0.8 &	0.3617 & 0.4314\\
    Dropwave & 0.6 & 0.0455 & 0.0452 \\
    \bottomrule
  \end{tabular}
  }
  \end{small}
\end{table*}

\textbf{Additional Optimization Tasks.} In Table~\ref{table:additional-benchmarks}, we evaluate our method on seven benchmark optimization tasks \citep{simulationlib} from the Bayesian optimization literature. We fixed the choice of hyperparameters in the calibration algorithm (3 randomly chosen initialization points, Matern kernel, PI acquisition, time-series splits and 25 BO steps). Similar to Table~\ref{table:baselines}, the error bars on reported AUC values are $<0.02$ and $f$ is not amenable to error bar calculation. 

On some functions (e.g., Dropwave, Beal), the calibrated and uncalibrated methods perform similarly. The Cross-in-Tray function has multiple sharp edges, possibly making it harder for the GP to model and worsening the performance of our method. On the McCormick function, the uncalibrated method makes more progress in the initial stages, but the calibrated method finds the minimum earlier.

\textbf{Sensitivity Analysis.} We perform a sensitivity analysis of our method over six of its hyper-parameters (e.g., kernel types, acquisition functions) for the Forrester function. 
 See Table~\ref{table:sensitivity-analysis} in Appendix~\ref{apdx:sensitivity-analysis} for the full experimental results. 
 Our results indicate good robustness across hyper-parameters; we also show slight improvements from our cross-validation heuristic.
\subsection{Hyperparameter Optimization Tasks}



\begin{figure*}[h!]
\centering     
\subfigure[Online LDA]{\label{fig:onlinelda}\includegraphics[width=0.3\linewidth]{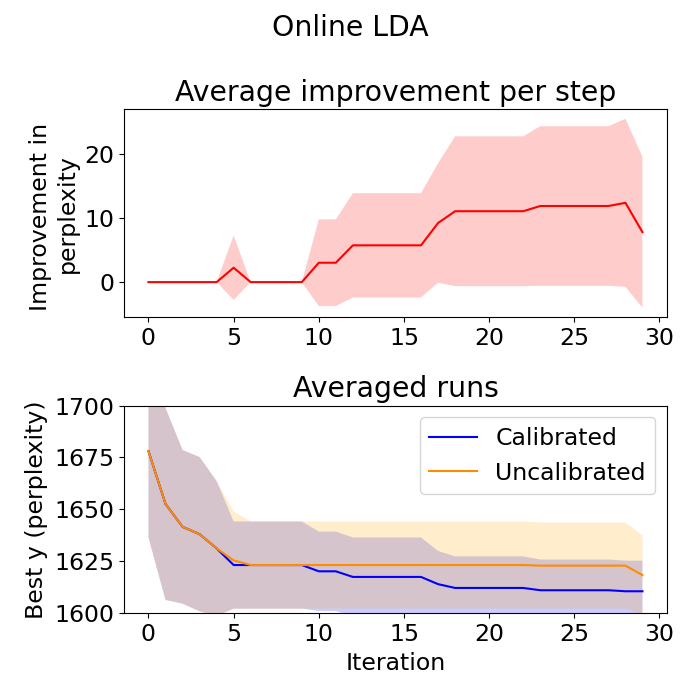}}
\subfigure[CIFAR10]{\label{fig:cifar10}\includegraphics[width=0.3\linewidth]{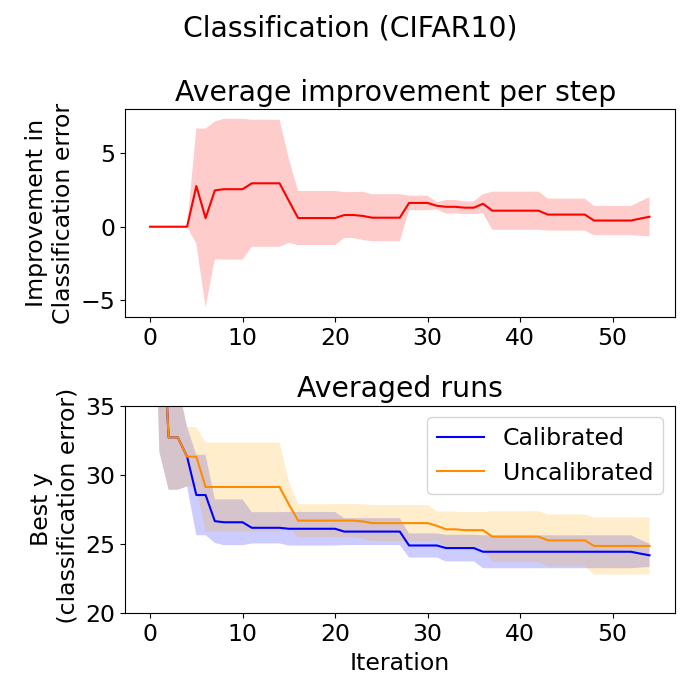}}
\subfigure[SVHN]{\label{fig:svhn}\includegraphics[width=0.3\linewidth]{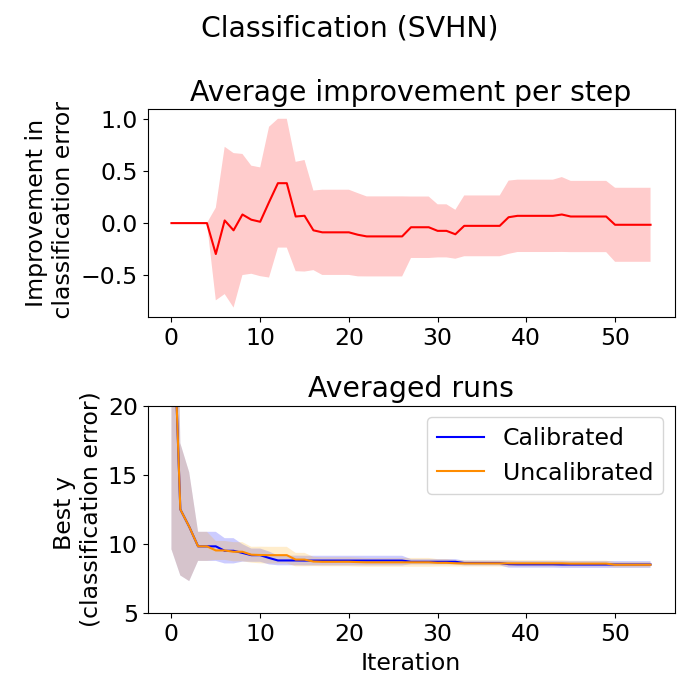}}
\caption{
Hyperparameter Optimization Experiments. 
\textbf{Top:} The average improvement made by the calibrated method over the uncalibrated method. 
\textbf{Bottom:} The best minimum found by each method per iteration. 
}
\vspace{10pt}
\end{figure*}

\paragraph{Online LDA}
In the Online LDA algorithm \citep{hoffman_NIPS2010_71f6278d}, we have three hyperparameters: $\tau_0$ and $\kappa$ which control the learning rate and minibatch size $s$ (Appendix~\ref{apdx:online_lda}). 
The objective function runs the Online LDA algorithm with these hyperparameters to convergence on the training set and outputs test set perplexity. We run this experiment on the 20 Newsgroups dataset. 
In Figure~\ref{fig:onlinelda}, we see that the calibrated method achieves a configuration of hyperparameters giving lower average perplexity. The error bars around the averaged runs are intersecting significantly due to variation across experiment repetitions. Hence, we add a separate plot showing the average improvement over time, defined as the difference between the best minimum found by the uncalibrated method and the best minimum found by the calibrated method. We observe that it is positive most of the time. 

\paragraph{Image Classification}

We define a Convolutional Neural Network (CNN) for image classification with 6 tunable hyperparameters: batch size, learning rate, filter size, etc. (see Appendix~\ref{apdx:image_classification}). Our objective function trains the CNN and returns the classification error on test dataset. 
We run our experiments on CIFAR10 \citep{Krizhevsky09learningmultiple} and SVHN datasets \citep{svhndataset}.  On the CIFAR10 dataset (Figure ~\ref{fig:cifar10}), we see that the calibrated method achieves a lower classification error on an average at the end of 50 iterations of Bayesian optimization. The calibrated method also achieves this minimum using about 50\% less number of steps.  
\newpage
\newpage

\section{DISCUSSION \& RELATED WORK}

Bayesian optimization is commonly used for optimizing black-box objective functions in applications like robotics \citep{calandra2016bayesian}, reinforcement learning \citep{brochu2010tutorial}, hyperparameter optimization \citep{Bergstra_NIPS2011_HPOAlgs}, recommender systems   \citep{Vanchinathan2014ExploreexploitIT}, automatic machine learning,  \citep{thornton2013autoweka} and materials design \citep{Frazier_2015}. The choices like acquisition function \citep{snoek2012practicalbo}, kernel function \citep{duvenaud2013structure} and transformation of input spaces \citep{snoek2014input} are important in making sure that the optimization works efficiently. 
Calibration in online sequential decision making and iterative design tasks has also been explored for prediction sets from a conformal prediction perspective
\citep{fannjiang2022design, stanton2023bayesian}.
While existing works approach a similar problem, they address distribution mismatch via techniques based on importance sampling, while our paper develops techniques based on online learning.

\paragraph{Calibrated Uncertainties}
Platt scaling \citep{Platt99probabilisticoutputs} and isotonic regression \citep{mizil2005predicting} are popular ways for calibrating uncertainties. This concept can be extended to regression calibration \citep{kuleshov2018accurate}, distribution calibration \citep{song2019distribution}, online learning \citep{kuleshov2017estimating}, and structured prediction \citep{kuleshov2015structured}.
Accurate uncertainty representation has been studied for applications like model-based reinforcement learning \mbox{\citep{malik2019calibrated}}, semi-supervised learning \citep{kuleshov2017deep}, neural networks \citep{guo2017calibration} and natural language processing models \mbox{\citep{nguyen2015posterior}}. 

\paragraph{Limitations}
There are still some tasks such as SVHN image classification where calibration does not improve performance. Studying the properties of objective functions where calibrated Bayesian optimization produces lower benefits can be useful in identifying directions for further improvements in the algorithm.


\paragraph{Conclusion}

The accuracy of uncertainty envelopes is important for balancing exploration and exploitation in Bayesian optimization. We show that we can calibrate the probabilistic model without additional function evaluations. Our approach improves the performance of Bayesian optimization in standard benchmark functions and hyperparameter optimization tasks.

\newpage

\bibliography{bibliography}

\begin{thebibliography}{45}
\providecommand{\natexlab}[1]{#1}
\providecommand{\url}[1]{\texttt{#1}}
\expandafter\ifx\csname urlstyle\endcsname\relax
  \providecommand{\doi}[1]{doi: #1}\else
  \providecommand{\doi}{doi: \begingroup \urlstyle{rm}\Url}\fi

\bibitem[authors(2016)]{gpyopt2016}
The~GPyOpt authors.
\newblock Gpyopt: A bayesian optimization framework in python.
\newblock \url{http://github.com/SheffieldML/GPyOpt}, 2016.

\bibitem[Bergstra et~al.(2011)Bergstra, Bardenet, Bengio, and K\'{e}gl]{Bergstra_NIPS2011_HPOAlgs}
James Bergstra, R\'{e}mi Bardenet, Yoshua Bengio, and Bal\'{a}zs K\'{e}gl.
\newblock Algorithms for hyper-parameter optimization.
\newblock In J.~Shawe-Taylor, R.~Zemel, P.~Bartlett, F.~Pereira, and K.~Q. Weinberger, editors, \emph{Advances in Neural Information Processing Systems}, volume~24. Curran Associates, Inc., 2011.
\newblock URL \url{https://proceedings.neurips.cc/paper/2011/file/86e8f7ab32cfd12577bc2619bc635690-Paper.pdf}.

\bibitem[Berkenkamp et~al.(2019)Berkenkamp, Schoellig, and Krause]{berkenkamp2019noregret}
Felix Berkenkamp, Angela~P. Schoellig, and Andreas Krause.
\newblock No-regret bayesian optimization with unknown hyperparameters, 2019.

\bibitem[Brochu et~al.(2010)Brochu, Cora, and de~Freitas]{brochu2010tutorial}
Eric Brochu, Vlad~M. Cora, and Nando de~Freitas.
\newblock A tutorial on bayesian optimization of expensive cost functions, with application to active user modeling and hierarchical reinforcement learning, 2010.

\bibitem[Calandra et~al.(2016)Calandra, Seyfarth, and Peters]{calandra2016bayesian}
R.~Calandra, A.~Seyfarth, and J~Peters.
\newblock Bayesian optimization for learning gaits under uncertainty.
\newblock \emph{Ann Math Artif Intell}, page 5–23, 2016.
\newblock URL \url{https://doi.org/10.1007/s10472-015-9463-9}.

\bibitem[Duvenaud et~al.(2013)Duvenaud, Lloyd, Grosse, Tenenbaum, and Ghahramani]{duvenaud2013structure}
David Duvenaud, James~Robert Lloyd, Roger Grosse, Joshua~B. Tenenbaum, and Zoubin Ghahramani.
\newblock Structure discovery in nonparametric regression through compositional kernel search, 2013.

\bibitem[Fannjiang et~al.(2022)Fannjiang, Bates, Angelopoulos, Listgarten, and Jordan]{fannjiang2022design}
Clara Fannjiang, Stephen Bates, Anastasios~N. Angelopoulos, Jennifer Listgarten, and Michael~I. Jordan.
\newblock Conformal prediction for the design problem.
\newblock \emph{CoRR}, abs/2202.03613, 2022.
\newblock URL \url{https://arxiv.org/abs/2202.03613}.

\bibitem[Frazier(2018)]{frazier2018tutorial}
Peter~I. Frazier.
\newblock A tutorial on bayesian optimization, 2018.

\bibitem[Frazier and Wang(2015)]{Frazier_2015}
Peter~I. Frazier and Jialei Wang.
\newblock Bayesian optimization for materials design.
\newblock \emph{Springer Series in Materials Science}, page 45–75, Dec 2015.
\newblock ISSN 2196-2812.
\newblock \doi{10.1007/978-3-319-23871-5_3}.
\newblock URL \url{http://dx.doi.org/10.1007/978-3-319-23871-5_3}.

\bibitem[Gibbs and Candes(2021)]{gibbs2021adaptive}
Isaac Gibbs and Emmanuel Candes.
\newblock Adaptive conformal inference under distribution shift.
\newblock \emph{Advances in Neural Information Processing Systems}, 34:\penalty0 1660--1672, 2021.

\bibitem[Gneiting and Raftery(2007)]{gneiting2007strictly}
Tilmann Gneiting and Adrian~E Raftery.
\newblock Strictly proper scoring rules, prediction, and estimation.
\newblock \emph{Journal of the American Statistical Association}, 102\penalty0 (477):\penalty0 359--378, 2007.
\newblock \doi{10.1198/016214506000001437}.
\newblock URL \url{https://doi.org/10.1198/016214506000001437}.

\bibitem[Guo et~al.(2017)Guo, Pleiss, Sun, and Weinberger]{guo2017calibration}
Chuan Guo, Geoff Pleiss, Yu~Sun, and Kilian~Q. Weinberger.
\newblock On calibration of modern neural networks, 2017.

\bibitem[Hazan et~al.(2016)]{hazan2016introduction}
Elad Hazan et~al.
\newblock Introduction to online convex optimization.
\newblock \emph{Foundations and Trends{\textregistered} in Optimization}, 2\penalty0 (3-4):\penalty0 157--325, 2016.

\bibitem[Hoffman et~al.(2010)Hoffman, Bach, and Blei]{hoffman_NIPS2010_71f6278d}
Matthew Hoffman, Francis Bach, and David Blei.
\newblock Online learning for latent dirichlet allocation.
\newblock In J.~Lafferty, C.~Williams, J.~Shawe-Taylor, R.~Zemel, and A.~Culotta, editors, \emph{Advances in Neural Information Processing Systems}, volume~23. Curran Associates, Inc., 2010.
\newblock URL \url{https://proceedings.neurips.cc/paper/2010/file/71f6278d140af599e06ad9bf1ba03cb0-Paper.pdf}.

\bibitem[Koenker and Bassett~Jr(1978)]{koenker1978regression}
Roger Koenker and Gilbert Bassett~Jr.
\newblock Regression quantiles.
\newblock \emph{Econometrica: journal of the Econometric Society}, pages 33--50, 1978.

\bibitem[Krizhevsky(2009)]{Krizhevsky09learningmultiple}
Alex Krizhevsky.
\newblock Learning multiple layers of features from tiny images.
\newblock Technical report, 2009.

\bibitem[Kuleshov and Ermon(2017{\natexlab{a}})]{kuleshov2017deep}
Volodymyr Kuleshov and Stefano Ermon.
\newblock Deep hybrid models: bridging discriminative and generative approaches.
\newblock In \emph{Uncertainty in Artificial Intelligence}, 2017{\natexlab{a}}.

\bibitem[Kuleshov and Ermon(2017{\natexlab{b}})]{kuleshov2017estimating}
Volodymyr Kuleshov and Stefano Ermon.
\newblock Estimating uncertainty online against an adversary.
\newblock In \emph{AAAI}, pages 2110--2116, 2017{\natexlab{b}}.

\bibitem[Kuleshov and Liang(2015)]{kuleshov2015structured}
Volodymyr Kuleshov and Percy~S Liang.
\newblock Calibrated structured prediction.
\newblock In C.~Cortes, N.~Lawrence, D.~Lee, M.~Sugiyama, and R.~Garnett, editors, \emph{Advances in Neural Information Processing Systems}, volume~28. Curran Associates, Inc., 2015.
\newblock URL \url{https://proceedings.neurips.cc/paper/2015/file/52d2752b150f9c35ccb6869cbf074e48-Paper.pdf}.

\bibitem[Kuleshov et~al.(2018)Kuleshov, Fenner, and Ermon]{kuleshov2018accurate}
Volodymyr Kuleshov, Nathan Fenner, and Stefano Ermon.
\newblock Accurate uncertainties for deep learning using calibrated regression, 2018.

\bibitem[Lakshminarayanan et~al.(2016)Lakshminarayanan, Pritzel, and Blundell]{lakshmi2016simple}
Balaji Lakshminarayanan, Alexander Pritzel, and Charles Blundell.
\newblock Simple and scalable predictive uncertainty estimation using deep ensembles, 2016.
\newblock URL \url{https://arxiv.org/abs/1612.01474}.

\bibitem[Malik et~al.(2019)Malik, Kuleshov, Song, Nemer, Seymour, and Ermon]{malik2019calibrated}
Ali Malik, Volodymyr Kuleshov, Jiaming Song, Danny Nemer, Harlan Seymour, and Stefano Ermon.
\newblock Calibrated model-based deep reinforcement learning, 2019.

\bibitem[Murphy and Winkler(1987)]{murphy1987general}
Allan~H Murphy and Robert~L Winkler.
\newblock A general framework for forecast verification.
\newblock \emph{Monthly weather review}, 115\penalty0 (7):\penalty0 1330--1338, 1987.

\bibitem[Netzer et~al.(2011)Netzer, Wang, Coates, Bissacco, Wu, and Ng]{svhndataset}
Yuval Netzer, Tao Wang, Adam Coates, Alessandro Bissacco, Bo~Wu, and Andrew~Y. Ng.
\newblock Reading digits in natural images with unsupervised feature learning.
\newblock In \emph{NIPS Workshop on Deep Learning and Unsupervised Feature Learning 2011}, 2011.
\newblock URL \url{http://ufldl.stanford.edu/housenumbers/nips2011_housenumbers.pdf}.

\bibitem[Nguyen and O'Connor(2015)]{nguyen2015posterior}
Khanh Nguyen and Brendan O'Connor.
\newblock Posterior calibration and exploratory analysis for natural language processing models, 2015.

\bibitem[Niculescu-Mizil and Caruana(2005)]{mizil2005predicting}
Alexandru Niculescu-Mizil and Rich Caruana.
\newblock Predicting good probabilities with supervised learning.
\newblock In \emph{Proceedings of the 22nd International Conference on Machine Learning}, ICML '05, page 625–632, New York, NY, USA, 2005. Association for Computing Machinery.
\newblock ISBN 1595931805.
\newblock \doi{10.1145/1102351.1102430}.
\newblock URL \url{https://doi.org/10.1145/1102351.1102430}.

\bibitem[Paciorek and Schervish(2003)]{paciorek2003nonstationary}
Christopher Paciorek and Mark Schervish.
\newblock Nonstationary covariance functions for gaussian process regression.
\newblock In S.~Thrun, L.~Saul, and B.~Sch\"{o}lkopf, editors, \emph{Advances in Neural Information Processing Systems}, volume~16. MIT Press, 2003.
\newblock URL \url{https://proceedings.neurips.cc/paper_files/paper/2003/file/326a8c055c0d04f5b06544665d8bb3ea-Paper.pdf}.

\bibitem[Platt(1999)]{Platt99probabilisticoutputs}
John~C. Platt.
\newblock Probabilistic outputs for support vector machines and comparisons to regularized likelihood methods.
\newblock In \emph{ADVANCES IN LARGE MARGIN CLASSIFIERS}, pages 61--74. MIT Press, 1999.

\bibitem[Rasmussen and Williams(2005)]{Rasmussen2005GPML}
Carl~Edward Rasmussen and Christopher K.~I. Williams.
\newblock \emph{Gaussian Processes for Machine Learning (Adaptive Computation and Machine Learning)}.
\newblock The MIT Press, 2005.
\newblock ISBN 026218253X.

\bibitem[Rios and Tobar(2018)]{rios2018}
Gonzalo Rios and Felipe Tobar.
\newblock Learning non-gaussian time series using the box-cox gaussian process.
\newblock pages 1--8, 07 2018.
\newblock \doi{10.1109/IJCNN.2018.8489648}.

\bibitem[Shafer and Vovk(2007)]{tutorialconformal2007shafer}
Glenn Shafer and Vladimir Vovk.
\newblock A tutorial on conformal prediction.
\newblock 2007.
\newblock \doi{10.48550/ARXIV.0706.3188}.
\newblock URL \url{https://arxiv.org/abs/0706.3188}.

\bibitem[Shahriari et~al.(2016)Shahriari, Swersky, Wang, Adams, and de~Freitas]{Shahriari2016BOReview}
Bobak Shahriari, Kevin Swersky, Ziyu Wang, Ryan~P. Adams, and Nando de~Freitas.
\newblock Taking the human out of the loop: A review of bayesian optimization.
\newblock \emph{Proceedings of the IEEE}, 104\penalty0 (1):\penalty0 148--175, 2016.
\newblock \doi{10.1109/JPROC.2015.2494218}.

\bibitem[Shalev-Shwartz et~al.(2012)]{shalev2012online}
Shai Shalev-Shwartz et~al.
\newblock Online learning and online convex optimization.
\newblock \emph{Foundations and Trends{\textregistered} in Machine Learning}, 4\penalty0 (2):\penalty0 107--194, 2012.

\bibitem[Snelson et~al.(2004)Snelson, Ghahramani, and Rasmussen]{NIPS2003_6b5754d7}
Edward Snelson, Zoubin Ghahramani, and Carl Rasmussen.
\newblock Warped gaussian processes.
\newblock In S.~Thrun, L.~Saul, and B.~Sch\"{o}lkopf, editors, \emph{Advances in Neural Information Processing Systems}, volume~16. MIT Press, 2004.
\newblock URL \url{https://proceedings.neurips.cc/paper/2003/file/6b5754d737784b51ec5075c0dc437bf0-Paper.pdf}.

\bibitem[Snoek et~al.(2012)Snoek, Larochelle, and Adams]{snoek2012practicalbo}
Jasper Snoek, Hugo Larochelle, and Ryan~P. Adams.
\newblock Practical bayesian optimization of machine learning algorithms.
\newblock In \emph{Proceedings of the 25th International Conference on Neural Information Processing Systems - Volume 2}, NIPS'12, page 2951–2959, Red Hook, NY, USA, 2012. Curran Associates Inc.

\bibitem[Snoek et~al.(2014)Snoek, Swersky, Zemel, and Adams]{snoek2014input}
Jasper Snoek, Kevin Swersky, Richard~S. Zemel, and Ryan~P. Adams.
\newblock Input warping for bayesian optimization of non-stationary functions, 2014.

\bibitem[Snoek et~al.(2015)Snoek, Rippel, Swersky, Kiros, Satish, Sundaram, Patwary, Prabhat, and Adams]{pmlr-v37-snoek15}
Jasper Snoek, Oren Rippel, Kevin Swersky, Ryan Kiros, Nadathur Satish, Narayanan Sundaram, Mostofa Patwary, Mr~Prabhat, and Ryan Adams.
\newblock Scalable bayesian optimization using deep neural networks.
\newblock In Francis Bach and David Blei, editors, \emph{Proceedings of the 32nd International Conference on Machine Learning}, volume~37 of \emph{Proceedings of Machine Learning Research}, pages 2171--2180, Lille, France, 07--09 Jul 2015. PMLR.
\newblock URL \url{http://proceedings.mlr.press/v37/snoek15.html}.

\bibitem[Song et~al.(2019)Song, Diethe, Kull, and Flach]{song2019distribution}
Hao Song, Tom Diethe, Meelis Kull, and Peter Flach.
\newblock Distribution calibration for regression, 2019.

\bibitem[Springenberg et~al.(2016)Springenberg, Klein, Falkner, and Hutter]{NIPS2016_a96d3afe}
Jost~Tobias Springenberg, Aaron Klein, Stefan Falkner, and Frank Hutter.
\newblock Bayesian optimization with robust bayesian neural networks.
\newblock In D.~Lee, M.~Sugiyama, U.~Luxburg, I.~Guyon, and R.~Garnett, editors, \emph{Advances in Neural Information Processing Systems}, volume~29. Curran Associates, Inc., 2016.
\newblock URL \url{https://proceedings.neurips.cc/paper/2016/file/a96d3afec184766bfeca7a9f989fc7e7-Paper.pdf}.

\bibitem[Stanton et~al.(2023)Stanton, Maddox, and Wilson]{stanton2023bayesian}
Samuel Stanton, Wesley Maddox, and Andrew~Gordon Wilson.
\newblock Bayesian optimization with conformal prediction sets.
\newblock In \emph{International Conference on Artificial Intelligence and Statistics}, pages 959--986. PMLR, 2023.

\bibitem[Surjanovic and Bingham()]{simulationlib}
S.~Surjanovic and D.~Bingham.
\newblock Virtual library of simulation experiments: Test functions and datasets.
\newblock Retrieved October 8, 2022, from \url{http://www.sfu.ca/~ssurjano}.

\bibitem[Thornton et~al.(2013)Thornton, Hutter, Hoos, and Leyton-Brown]{thornton2013autoweka}
Chris Thornton, Frank Hutter, Holger~H. Hoos, and Kevin Leyton-Brown.
\newblock Auto-weka: Combined selection and hyperparameter optimization of classification algorithms, 2013.

\bibitem[Vanchinathan et~al.(2014)Vanchinathan, Nikolic, Bona, and Krause]{Vanchinathan2014ExploreexploitIT}
Hastagiri~P. Vanchinathan, I.~Nikolic, F.~D. Bona, and Andreas Krause.
\newblock Explore-exploit in top-n recommender systems via gaussian processes.
\newblock In \emph{RecSys '14}, 2014.

\bibitem[Vovk et~al.(2020)Vovk, Petej, Toccaceli, Gammerman, Ahlberg, and Carlsson]{VovkPTGAC20}
Vladimir Vovk, Ivan Petej, Paolo Toccaceli, Alexander Gammerman, Ernst Ahlberg, and Lars Carlsson.
\newblock Conformal calibrators.
\newblock In Alexander Gammerman, Vladimir Vovk, Zhiyuan Luo, Evgueni~N. Smirnov, Giovanni Cherubin, and Marco Christini, editors, \emph{Conformal and Probabilistic Prediction and Applications, {COPA} 2020, 9-11 September 2020, Virtual Event, Verona, Italy}, volume 128 of \emph{Proceedings of Machine Learning Research}, pages 84--99. {PMLR}, 2020.
\newblock URL \url{http://proceedings.mlr.press/v128/vovk20a.html}.

\bibitem[Zhao et~al.(2020)Zhao, Ma, and Ermon]{zhao2020individual}
Shengjia Zhao, Tengyu Ma, and Stefano Ermon.
\newblock Individual calibration with randomized forecasting, 2020.
\newblock URL \url{https://arxiv.org/abs/2006.10288}.

\end{thebibliography}

\newpage

\section*{Checklist}

 \begin{enumerate}

 \item For all models and algorithms presented, check if you include:
 \begin{enumerate}
   \item A clear description of the mathematical setting, assumptions, algorithm, and/or model. [Yes]
   \item An analysis of the properties and complexity (time, space, sample size) of any algorithm. [Yes]
   \item (Optional) Anonymized source code, with specification of all dependencies, including external libraries. [Yes]
 \end{enumerate}

 \item For any theoretical claim, check if you include:
 \begin{enumerate}
   \item Statements of the full set of assumptions of all theoretical results. [Yes]
   \item Complete proofs of all theoretical results. [Yes]
   \item Clear explanations of any assumptions. [Yes]     
 \end{enumerate}

 \item For all figures and tables that present empirical results, check if you include:
 \begin{enumerate}
   \item The code, data, and instructions needed to reproduce the main experimental results (either in the supplemental material or as a URL). [No]
   \item All the training details (e.g., data splits, hyperparameters, how they were chosen). [Yes]
         \item A clear definition of the specific measure or statistics and error bars (e.g., with respect to the random seed after running experiments multiple times). [Yes]
         \item A description of the computing infrastructure used. (e.g., type of GPUs, internal cluster, or cloud provider). [Yes]
 \end{enumerate}

 \item If you are using existing assets (e.g., code, data, models) or curating/releasing new assets, check if you include:
 \begin{enumerate}
   \item Citations of the creator If your work uses existing assets. [Yes]
   \item The license information of the assets, if applicable. [Not Applicable]
   \item New assets either in the supplemental material or as a URL, if applicable. [Not Applicable]
   \item Information about consent from data providers/curators. [Not Applicable]
   \item Discussion of sensible content if applicable, e.g., personally identifiable information or offensive content. [Not Applicable]
 \end{enumerate}

 \item If you used crowdsourcing or conducted research with human subjects, check if you include:
 \begin{enumerate}
   \item The full text of instructions given to participants and screenshots. [Not Applicable]
   \item Descriptions of potential participant risks, with links to Institutional Review Board (IRB) approvals if applicable. [Not Applicable]
   \item The estimated hourly wage paid to participants and the total amount spent on participant compensation. [Not Applicable]
 \end{enumerate}

 \end{enumerate}

\newpage
\onecolumn

\appendix
\onecolumn
\aistatstitle{Online Calibrated Uncertainty Estimation for Bayesian Optimization (Supplementary Material)}

\section{ADDITIONAL DETAILS ON EXPERIMENTS}
\label{apdx:experimental_details}
 We implement our method on top of the GPyOpt library ~\citep{gpyopt2016} (BSD 3 Clause License) in Python. The output values of objective function are normalized before training the base GP model. The GP uses Radial Basis Function (RBF) kernel. For the experiments on hyperparameter optimization tasks, we used Expected Improvement as acquisition function. We use 5 randomly chosen data-points to initialize the base GP. The experiments are repeated 5 times and the results are averaged over these 5 runs. 
 
\begin{table*}[!h]
  \caption{Comparison of Calibrated Bayesian Optimization Against Baselines. We see that the calibrated method compares favorably against Input warping (\citep{snoek2014input}),  Output warping (tanh as in \citep{NIPS2003_6b5754d7}), Boxcox and Log transformation of output (\citep{rios2018}), and Bagged ensemble of 5 GPs (\citep{lakshmi2016simple}).}
  \label{table:baselines-apdx}
  \centering
  \resizebox{\textwidth}{!}{
  
  \begin{tabular}{lllcc}
    
    Objective & Method	& Minimum Found & Fraction of Runs Where  & 	Area Under the Curve \\
    	Function & & (↓) & Calibrated Beats Baseline (↑) & 	 (↓)\\
    \midrule
    
    Forrester (1D) 
    
    &Calibrated Method & -4.983 (0.894)&	1	& 0.8187\\
    &Uncalibrated method  & -0.986 (0.001) &	0.8 & 0.9866\\
    &Input warping & -0.889 (0.000) &1& 1.0000\\
    &Output warping & -6.021 (0.000) & 0.0 & 0.2409\\
    &Boxcox transformation of output & -4.806 (0.780) & 0.4 & 0.4911\\
    &Log transformation of output & -0.986 (0.000) & 1 & 0.9865\\
    &Bagged ensemble of 5 GPs & -3.891(1.098) & 0.8& 0.8382\\

    \midrule

    Ackley (2D) 
    &Calibrated Method & 5.998 (2.314) & 1 &	0.5516\\
    &Uncalibrated method    & 12.359 (2.439) & 0.8	& 0.7815\\
    &Input warping  & 14.061 (1.312) & 0.8	& 0.8125\\
    &Output warping      & 10.459 (3.365) & 0.6 &	0.7257     \\
      
    &Boxcox transformation of output   & 14.421 (1.423)&	0.8	&0.9015     \\
    &Log transformation of output & 8.330 (0.849)&	0.8	& 0.6524     \\
    &Bagged ensemble of 5 GPs    & 6.011 (2.544)	& 0.6	& 0.6013   \\
    \midrule

    Cosines (2D) 
    &Calibrated Method & -1.5983 (0.0006) &	1 &	0.2969\\
    &Uncalibrated method  &-1.5974 (0.00102)	& 0.8	& 0.3441\\
    &Input warping  & -1.3054 (0.0347)	& 1.0	& 0.9534\\
    &Output warping  & -1.5994 (0.0001)	& 0.8 &	0.4730\\
    &Boxcox transformation of output & -1.5306 (0.0590) &	0.8	& 0.5969\\
    &Log transformation of output & -1.5992 (0.0003)	& 0.8 &	0.3642\\
    &Bagged ensemble of 5 GPs  & -1.5989 (0.0003)	& 0.4 &	0.3179\\
        \midrule

    Alpine (10D)
    &Calibrated Method & 12.537 (0.909)	& 1	& 0.6423\\
    &Uncalibrated method  &15.506 (1.275) &	0.6 &	0.6527\\
    &Input warping  & 15.697 (1.740) &	0.8 &	0.6492\\
    &Output warping  & 13.531 (2.127)	& 0.6 &	0.5857\\
    &Boxcox transformation of output  & 15.715 (0.603)	& 0.8 &	0.6253\\
    &Log transformation of output  & 20.996 (1.661)	& 0.8 &	0.7931\\
    &Bagged ensemble of 5 GPs  & 16.677 (1.699)	& 0.8 &	0.7334\\
    \bottomrule
\end{tabular}
  }
\end{table*}

\subsection{Evaluation Metrics}
\label{apdx:evaluation_metrics}
We define the following metrics to compare calibrated Bayesian optimization with uncalibrated method and other baselines. 
\begin{enumerate}
    \item Minimum value $m$ of the objective function achieved by Bayesian optimization. Lower values of $m$ are better.
    \item Fraction $f$ of experimental repetitions where the baseline performs worse than our method. A baseline performs worse than our method if it does not find a lower minimum, if it finds the same minimum in a larger number of steps, or if its optimization curve is entirely above that of the calibrated method. Higher values of $f$ are better.
    \item Normalized area under the optimization curve \textbf{$a$}. For each method, we compute the area under its optimization curve (i.e., its optimum as a function of the number of optimization steps; see Figure~\ref{benchmark_comparison}) and normalize it relative to the area of the rectangle formed by upper and lower bounds along the $x$, $y$ axes (hence max \textbf{$a$} is one). Smaller values of \textbf{$a$} indicate that the method reaches the minimum faster, hence are better.
\end{enumerate}

\subsection{Comparing Against Additional Baselines}
\label{apdx:baseline-comparison}
 Our method outperformed the modern conformal baseline~\citep{stanton2023bayesian} that applies conformal prediction for calibrated uncertainties in Bayesian optimization. We also outperform the Adaptive UCB~\citep{berkenkamp2019noregret} method (that adjusts the UCB interval adaptively) on two tasks and are within the margin of error on the third. 

 We produce additional results with several other baselines in Table~\ref{table:baselines-apdx}.

\begin{figure}[h]
\vspace{-0.3cm}
\centering
  \includegraphics[scale=0.21]{figures/rebuttal/ackley_30.png}
  \includegraphics[scale=0.21]{figures/rebuttal/sixhumpcamel.png}
  \includegraphics[scale=0.21]{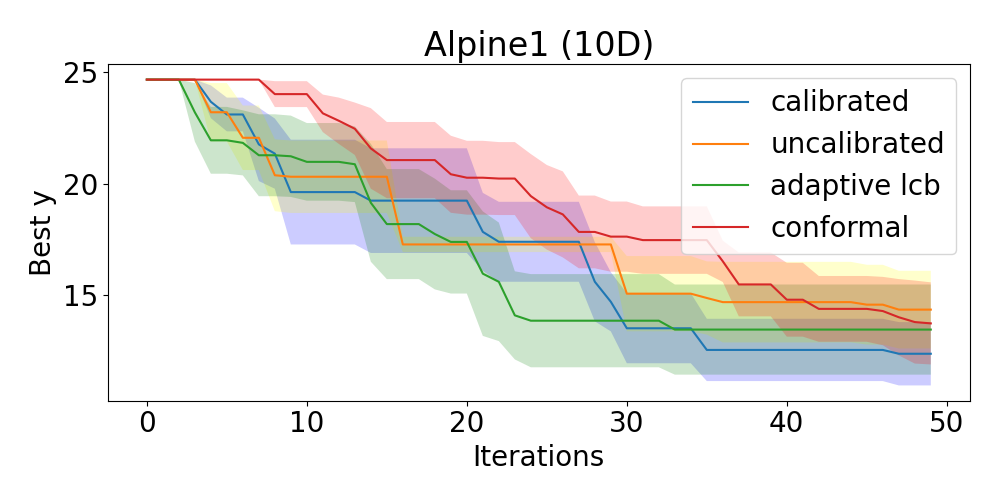}
  \label{fig:baselines}
  \vspace{-0.3cm}
\end{figure}

\subsection{Sensitivity Analysis}
\label{apdx:sensitivity-analysis}
We produce additional results on sensitivity analysis in Table~\ref{table:sensitivity-analysis}. 
\begin{table*}[htb]

  \caption{Sensitivity Analysis of Calibrated Bayesian Optimization (Algorithm~\ref{calibrated-bo} and ~\ref{alg:recal}) for the Forrester Function.}
  \label{table:sensitivity-analysis}
  \centering
  \resizebox{\textwidth}{!}{
  \begin{tabular}{l|lccc}
    Hyper-Parameter	& Modification &	\% Runs Where &	Area Under the Curve,  &	Area Under the Curve, \\
    	&  &	Calibrated is Best (↑) &	 Calibrated (↓) &	 Uncalibrated (↓)\\
    
    \midrule
   Kernel of Base Model $\mathcal{M}$& Matern	& 0.8	& 0.2811	& 0.3454 \\
& Linear & 	1.0 & 	0.5221	 & 1.0000\\
 & RBF	 & 0.4	 & 0.2366	 & 0.2922\\
 & Periodic & 	0.6 & 	0.2586 & 	0.3305\\
 \midrule
 Number of Time-series
  & N-1	 & 0.8	 & 0.2811	 & 0.3454\\
 splits in \textsc{CreateSplits} & N-2	 & 0.8	 & 0.2768	 & 0.3454\\
 & N-3	 & 0.8	 & 0.2653	 & 0.3454\\
 & N-4	 & 0.6	 & 0.2643	 & 0.3454\\
 \midrule
 Recalibrator Model $\mathcal{R}$
  & GP	 & 0.8	 & 0.2810 & 	0.3454\\
 & MLP	 & 0.8	 & 0.2785 & 	0.3454\\
 \midrule
Number of Data Points
 & 3	 & 0.8	 & 0.2810	 & 0.3454\\
 for Initializing Base Model & 4	 & 0.8	 & 0.0557	 & 0.0614\\
 & 7	 & 0.4	 & 0.0719	 & 0.0736\\
 & 10	 & 0.4	 & 0.0825	 & 0.0817\\
 \midrule
Initialization Design 
 & Random	 & 0.8	 & 0.0557	 & 0.0614\\
 (Base model initialized & Sobol	 & 0.6	 & 0.0415	 & 0.0414\\
  with 4  data points)& Latin	 & 0.2	 & 0.2358	 & 0.2181\\
 \midrule
Acquisition function 
 & LCB	 & 1.0	 & 1.0000	 & 1.0000\\
 (Linear kernel)& EI	 & 1.0	 & 0.5221	 & 1.0000\\
 & PI	 & 0.8	 & 0.6920	 & 1.0000\\
    \bottomrule
  \end{tabular}
  }
\end{table*}

\subsection{Increasing the number of optimization steps}
\label{apdx:increasing-optimization-step-num}



\begin{wrapfigure}{r}{0.35\textwidth}
\vspace{-7mm}
  \includegraphics[scale=0.22]{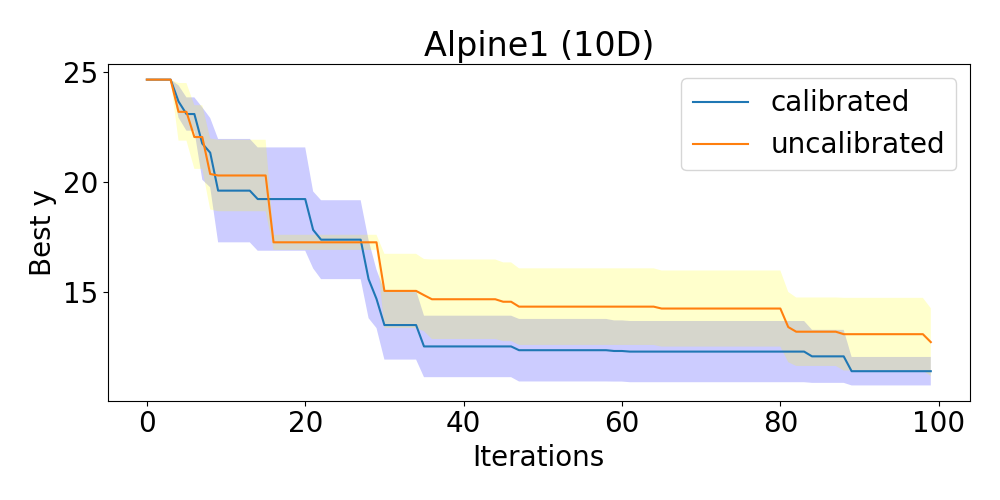}
  \label{fig:birds}
  \vspace{-7mm}
\end{wrapfigure}
On the most challenging benchmark function (Alpine 10D), \textbf{we ran our method for 100 steps and we observed results consistent with 25 steps} . We see an improvement in the minima found at the 100-th step from 12.75 to 11.43. Note that on non-synthetic hyperparameter optimization tasks, we ran for >50 steps and observed that the methods plateaued.
    

\subsection{Online LDA}
\label{apdx:online_lda}
We use the grid of parameters mentioned in Table~\ref{onlinelda-table} as the input domain while running Bayesian optimization. We run this algorithm on the 20 Newsgroups dataset which contains 20,000 news documents partitioned evenly across 20 different newsgroups. We train the algorithm on 11,000 randomly chosen documents. A test-dataset of 2200 articles is used to assess the perplexity. 
\begin{table}[!htb]
  \caption{Hyperparameters for Online LDA}
  \label{onlinelda-table}
  \centering
  \begin{tabular}{lll}
    Name of HP     & Bounds    & Type of domain \\
    \midrule
    Minibatch size & [1, 128] & Discrete (log-scale)\\
    $\kappa$    & [0.5, 1]  & Continuous (step-size=0.1) \\
    $\tau_0$     & [1, 32] & Discrete (log-scale)     \\
    \bottomrule
  \end{tabular}
\end{table}
\subsection{Image Classification Using Neural Networks}
\label{apdx:image_classification}
We provide the range of hyperparameters considered while performing Bayesian optimization to determine their optimal configuration with reference to the image classification experiments in Table~\ref{cifar10-table}. 
\begin{table}[!htb]
  \caption{Hyperparameters for CNN (CIFAR10 and SVHN classification)}
  \label{cifar10-table}
  \centering
  \begin{tabular}{lll}
    Name of HP     & Bounds    & Type of domain \\
    \midrule
    Batch size & [32, 512] & Discrete (step size 32)\\
    Learning rate    & [0.0000001, 0.1]  & Continuous (log-scale) \\
    Learning rate decay     & [0.0000001, 0.001] & Continuous (log-scale)     \\
    L2 regularization     & [0.0000001, 0.001] & Continuous (log-scale)     \\
    Outchannels in fc layer & [256, 512] & Discrete (step size=16) \\
    Outchannels in conv layer & [128, 256] & Discrete (step size=16) \\
    
    \bottomrule
  \end{tabular}
\end{table}

\section{CALIBRATION OF PROBABILISTIC MODEL}
\label{apdx:recalibration_algorithm}
For training a recalibrator over our probabilistic model, we compute the CDF $F_t$ at each data-point $y_t$ using the formulation $F_t=[\mathcal{M}(x_t)](y_t)$. This can be used to estimate the the empirical fraction of data-points below each quantile. Algorithm~\ref{train-recalib-general} based on based on \citet{kuleshov2018accurate} outlines this procedure.
\begin{algorithm}
  \caption{Calibration of Probabilistic Model 
  }
  \label{train-recalib-general}
  \textbf{Input:} Dataset of probabilistic forecasts and outcomes $\{[\mathcal{M}(x_t)](y_t), y_t\}_{t=1}^{N}$ 
  \begin{enumerate}
      \item Form recalibration set 
      $\mathcal{D} = \{[F_t, \hat{P}(F_t)\}_{t=1}^{N}$
      where $F_t=[\mathcal{M}(x_t)](y_t)$ and $\hat{P}(p) = |\{ y_t | [F_t\leq p, t=1,..,N\}|/N$.
      \item Train recalibrator model $\mathcal{R}$ on dataset $\mathcal{D}$.
  \end{enumerate}
\end{algorithm}

\section{EXAMINING AQUISITION FUNCTIONS}
\label{apdx:acquisition_functions}
We analyze the role of calibration in common acquisition functions used in Bayesian optimization.



\paragraph{Probability of Improvement.}

The probability of improvement is given by ${P(f(x) \geq (f(x^{+})+\epsilon)}$, where $\epsilon>0$ and $x^{+}$ is the previous best point. Note that this corresponds to ${1 - F_x(f(x^{+})+\epsilon)}$, where $F_x$ is the CDF at $x$ that is predicted by the model. In a quantile-calibrated model, these probabilities on average correspond to the empirical probability of observing an improvement event. This leads to acquisition function values that more accurately reflect the value of exploring specific regions.
Furthermore, if the model is calibrated, we keep working with calibrated values throughout the optimization process, as $x^{+}$ changes.

\paragraph{Expected Improvement.} The expected improvement can be defined as ${\mathbb{E}[\max(f(x) - f(x^{+}), 0)]}$. This corresponds to computing the expected value of the random variable ${R = \max(Y-c, 0)}$, where $Y$ is the random variable that we are trying to model by $\mathcal{M}$, and $c \in \mathbb{R}$ is a constant. If we have a calibrated distribution over $Y$, it is easy to derive from it a calibrated distribution over $R$. By Proposition \ref{prop:expectations}, we can estimate $\mathbb{E}[R]$ under the calibrated model, just as we can estimate the probability of improvement in expectation.

\paragraph{Upper Confidence Bounds.}

The UCB acquisition function for a Gaussian process is defined as ${\mu(x) + \gamma\cdot \sigma(x)}$ at point $x$. For non-Gaussian models, this naturally generalizes to a quantile $F_x^{-1}(\alpha)$ of the predicted distribution $F$. In this context, recalibration adjusts confidence intervals such that $\alpha \in [0,1]$ corresponds to an interval that is above the true $y$ a fraction $\alpha$ of the time. This makes it easier to select a hyper-parameter $\alpha$. Moreover, as $\alpha$ or $\gamma$ are typically annealed, calibration induces a better and smoother annealing schedule.

\section{ADDITIONAL DISCUSSION}
\label{apdx:Additional_discussion}
A key conceptual contribution of our work is a new angle for reasoning about uncertainty in the context of sequential decision-making. There exist many known decompositions of uncertainty, e.g. epistemic vs. aleatoric. Our work argues for using a different decomposition of uncertainty that is rarely used: calibration + sharpness.

This decomposition is interesting because the calibration property can be easily enforced in practice; at the same time this property greatly improves sequential decision-making for reasons we explain in the paper, and enforcing it results in significant practical benefits. This fact is currently underappreciated; our work contributes to a body of literature (see e.g., \citet{malik2019calibrated}) that helps popularize the idea of reasoning about uncertainty through the lens of calibration and sharpness, and can lead to significant practical improvements in uncertainty-aware algorithms that adopt our angle.

From a methodological perspective, our work resolves challenges in applying calibration in the context of Bayesian optimization. We introduce recalibration mechanisms based on leave-one-out cross-validation with a temporal ordering and we design specific classes of Gaussian recalibrators that are compatible with GP outputs and acquisition function inputs. We discovered that more naive applications of calibration fail, and our methods are non-trivial. Finally, we show empirically that our ideas have significant practical benefits.

\subsection{On the Computational Cost of Calibrated Bayesian Optimization}

Calibration increases the computational cost of Bayesian optimization (since we fit multiple GP models, and not just one). However, in most applications, we expect that the cost of fitting GPs will be negligible compared to the cost of evaluating the objective function at a datapoint. For example, in hyper-parameter optimization, the cost of training a new neural network with a new set of hyper-parameters vastly exceeds the cost of fitting a GP. Hence, calibrated and uncalibrated methods are in practice comparable in terms of their computational costs, and training multiple GPs does not limit the applicability of our method. 

In terms of time complexity, the increase in computational costs to run  Algorithm~\ref{alg:recal} after each standard Bayesian optimization step in Algorithm~\ref{calibrated-bo} depends linearly on the number of cross-validation splits ($|S|$) and the time complexity to train the model $\mathcal{M}$. The time complexity also depends on an additive term consisting of dataset size $N$ multiplied with the inference time complexity of model $\mathcal{M}$.  This additive term comes from running step 2 in Algorithm~\ref{alg:recal} cumulatively on all the $\mathcal{D}_{\text{test}}$ sets. An additive term also corresponds to time-complexity to train recalibrator $R$ in the end. The increase in overall space complexity depends linearly on the number of cross-validation splits $|S|$ and the size of dataset $N$ together with an additive term to store the trained recalibrator $\mathcal{R}$. In our experiments, the model $\mathcal{M}$ is itself a Gaussian Process, but other models can be also be used to perform calibrated Bayesian optimization. 

The experiments with real world hyperparameter optimization tasks were run on a GPU cluster since training the neural network and Online LDA models with a chosen set of hyperparameters incurs a high computational cost. However, all other experiments that compute blackbox objective function with analytic formulas could be performed on a laptop with 2.8GHz quad-core
Intel i7 processor.

\subsection{On Epistemic vs. Aleatoric Uncertainties}

Our method calibrates both epistemic and aleatoric uncertainties equally well. The concept of calibration is complementary and orthogonal to the concept of epistemic vs. aleatoric uncertainty. Our method takes any probabilistic prediction P(y) over y (regardless of whether uncertainties P(y) are epistemic or aleatoric) and recalibrates it, resulting in improved performance.

Specifically, let $\mathcal{M}(x)$ be a probabilistic model that outputs a probabilistic forecast $P(y)$ over the target y. The $\mathcal{M}(x)$ may model purely aleatoric uncertainties (e.g., $\mathcal{M}$ is a neural network with a softmax or Gaussian output layer) or epistemic uncertainties (e.g., $\mathcal{M}$ is a GP). In either case, $P(y)$ is just a distribution for which we can assess calibration. Our method improves calibration equally well regardless of the type of $\mathcal{M}(x)$ that generated $P(y)$. Improved calibration in turn increases optimization performance of both Bayesian and non-Bayesian base models $\mathcal{M}(x)$.
 For more details, please consider the analysis of Bayesian and non-Bayesian methods by ~\citet{kuleshov2018accurate} (our work extends their recalibration technique and inherits its properties).

\begin{figure*}[htb]
\centering

\subfigure[Uncalibrated Bayesian optimization produces overconfident intervals and the global minimum is not explored ]{\label{fig:plain-extended}\includegraphics[scale=0.25]{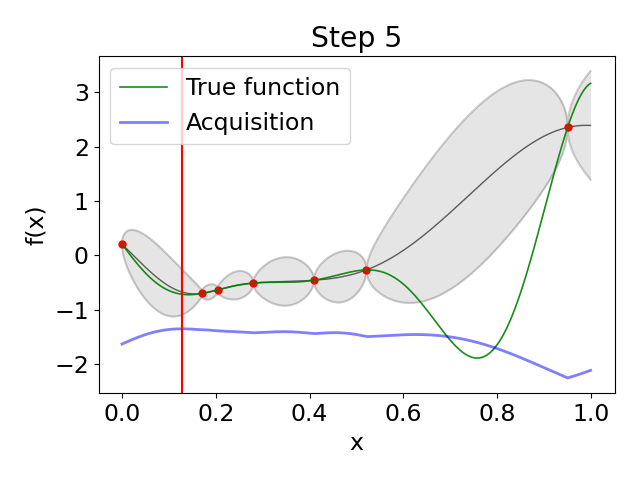}
\includegraphics[scale=0.25]{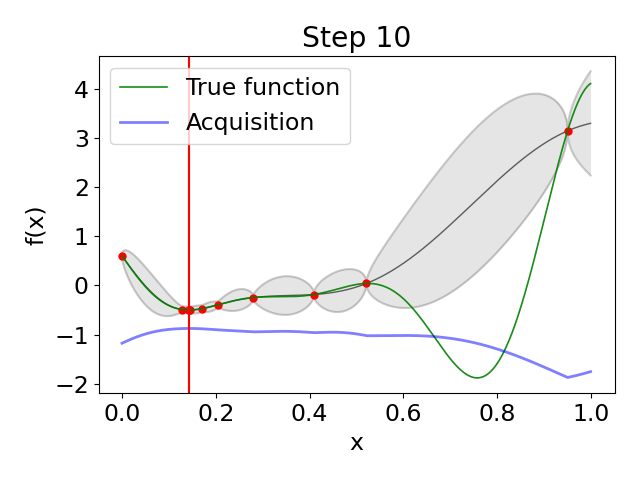}
\includegraphics[scale=0.25]{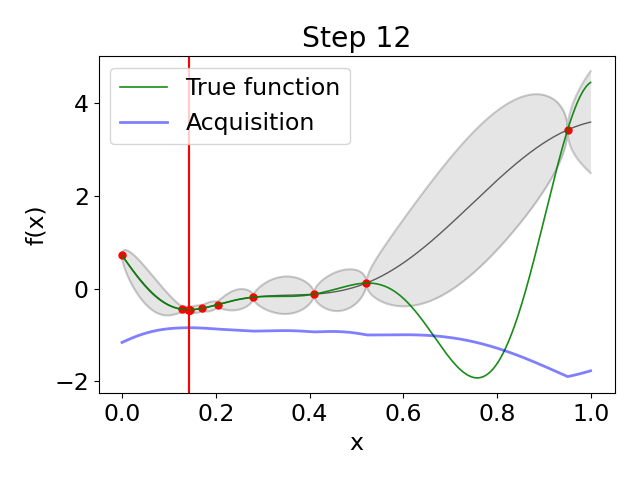}
\includegraphics[scale=0.25]{figures/forrestor_epochs/plain/epoch_14.png}
}

\subfigure[Calibrated Bayesian optimization produces wider confidence intervals at Step 12 and finds the global optimum]{\label{fig:calib-extended}
\includegraphics[scale=0.25]{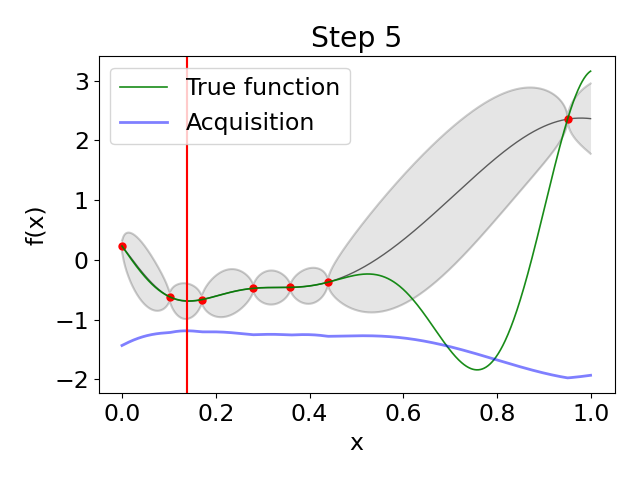}
\includegraphics[scale=0.25]{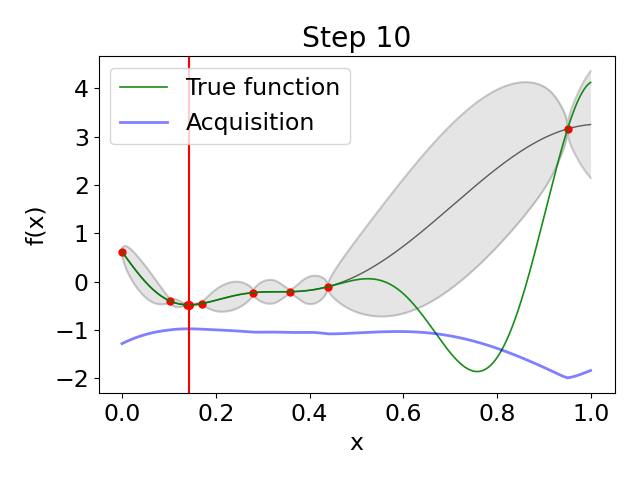}
\includegraphics[scale=0.25]{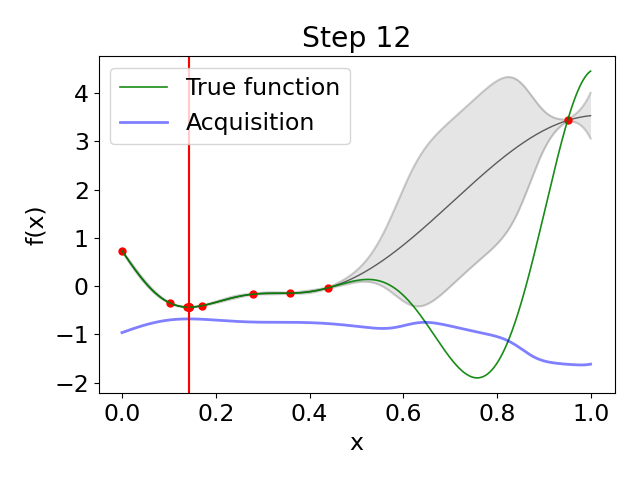}
\includegraphics[scale=0.25]{figures/forrestor_epochs/calib/epoch_14.png}}
\caption{Selected steps of uncalibrated and calibrated Bayesian optimization on the Forrester function (green) using the UCB acquisition function (blue). The global minimum lies near 0.8; however, after sampling 3 initial points at random, the model is constant in $[0.6, 0.9]$, while the true function has a large dip. Since confidence intervals in $[0.6, 0.9]$ are fairly narrow in the uncalibrated method, and the optimization algorithm never explores it, the global minimum is missed by a large margin. In the calibrated method however, the recalibrator learns after iteration 12 that the model is overconfident, expanding its confidence intervals. This leads the calibrated model to explore in $[0.6, 0.9]$ and find the global minimum. 
}
\label{fig:forrestor-epochs}
\end{figure*}
\subsection{On Sensitivity With Respect to the Gaussian Assumption} 
The effects of the Gaussian assumption can be seen in Table~\ref{table:sensitivity-analysis}. These additional tasks cover benchmark functions which are closer (e.g. cosines) and farther (e.g. cross-in-tray) from Gaussian assumptions. We do observe lower performance of both calibrated and uncalibrated Bayesian optimization methods on non-Gaussian tasks. This suggests an opportunity to improve Bayesian optimization by leveraging non-Gaussian models as a replacement for the classical GP approach.

\subsection{On Sensitivity With Respect to Higher Dimensions} 
In the paper, we have results for the Alpine function in 10 dimensions and the hyperparameter optimization tasks have 3-6 input dimensions each. Thus, we observe the benefits of calibration in higher dimensions as well. However, we did observe that in higher dimensions the improvement offered by calibration starts to become gradually less pronounced, which may be attributed to the curse of dimensionality (the difficulty of estimating densities in high dimensions).

\subsection{Non-stationarity in Bayesian Optimization and Calibration Literature}
\label{apdx:stationarity-clarification}
The notion of non-stationarity of outcome function in Bayesian optimization is different from non-stationarity of data distribution in calibration literature.

A stationary objective function in the context of Bayesian optimization refers to unchanging characteristics smoothness of the function with changing inputs~\citep{paciorek2003nonstationary}. 

For example, when modelled using a Gaussian Process (GP) regression,  stationarity of the modelled function refers to the property of translation invariance of covariance between two outputs~\citep{snoek2014input}. The properties of a GP regression are determined by the mean function $m: \mathcal{X} \rightarrow \mathbb{R}$ and covariance function or kernel $K: \mathcal{X} \times \mathcal{X} \rightarrow \mathbb{R}$. Given a set of data-points $\mathcal{D} = \{x_t, y_t\}_{t=1}^{T}$ such that $\mathbf{X} = \{x_t\}_{t=1}^{T}$ and $\mathbf{y} = \{ y_t\}_{t=1}^{T}$, the mean $\mu$ and covariance $\Sigma$ as modelled by the GP can be expressed as 

$$\mu({x}, \mathcal{D}) = m(\mathbf{X}) + K(\mathbf{X}, x)^T (K(\mathbf{X}, \mathbf{X})^{-1}) (\mathbf{y} - m(\mathbf{X}))$$

$$\Sigma(x, x', \mathcal{D}) = K(x, x') - K(\mathbf{X}, x)^T (K(\mathbf{X}, \mathbf{X})^{-1}) K(\mathbf{X}, x')$$

Here, an example of stationary kernel is the popular 5/2 Matern kernel

$K_{52}(x, x') = k \left(1 + \sqrt{5r^2} + \frac{5 r^2}{3}\right) \exp(-\sqrt{5r^2})$, where $r = \sum_{d=1}^{D} (x_d - x_d')^2$ for D-dimensional input. We can see that this kernel is invariant to translations in input space. If the smoothness of true output function varies with input (i.e., non-stationary function), then a stationary kernel in GP regression may not be adequate to model the function.

A non-stationary data-distribution in calibration, on the other hand, refers to the determination of next data-point dependent on previous data-points. The data-points chosen sequentially in Bayesian optimization are not independent of each other, thus producing a non-stationary data-distribution. 



\section{ANALYSING CALIBRATION FOR FORRESTER FUNCTION}
\label{apdx:visualize_forrester}
In Figure~\ref{fig:forrestor-epochs}, we see a visual comparison of optimization performed by calibrated method against the uncalibrated method.

\section{Monotonic Loss Bound}
\label{apdx:math_proofs}

We have shown that a calibrated model can be used to estimate expectations on average. Here, we complement these results with additional concentration inequalities which show that estimates of the calibrated loss do not exceed the true loss by too much. Note that this statement represents an extension of the Markov inequality.

\markovvariant*




\begin{proof}

Recall that $M(x)$ is a distribution over $\mathcal{Y}$, with a density $p_x$, a quantile function $Q_x$, and a cdf $F_x$.
Note that for any $x$ and $s \in (0,1)$ and $y' \leq F_x^{-1}(1-s)$ we have:
\begin{align*}
\ell(x)  
& = \int \ell(y, a(x), x) p_x(y) dy \\
& \geq \int_{y \geq y'} \ell( y, a(x), x) p_x(y) dy \\
& \geq \ell( y', a(x), x) \int_{y \geq y'} p_x(y) dy \\
& \geq s \ell(y', a(x), x)
\end{align*}

The above logic implies that whenever $\ell(x)   \leq s \ell(y, a, x)$, we have $y \geq F_x^{-1}(1-s)$ or $F_x(y) \geq (1-s)$. Thus, we have for all $t$,
\begin{align*}
\mathbb{I}\{  \ell(x_t)   \leq s \ell(y_t, a_t, x_t) \} \leq \mathbb{I}\{  F_{x_t}(y_t) \geq (1-s) \}.
\end{align*}
Therefore, we can write
\begin{align*}
\frac{1}{T} \sum_{t=1}^T \mathbb{I}\{  \ell(x_t)   \leq s \ell(y_t, a_t, x_t) \} \leq \frac{1}{T} \sum_{t=1}^T \mathbb{I}\{  F_{x_t}(y_t) \geq (1-s) \} = s + o(T),
\end{align*}
where the last equality follows because $M$ is calibrated. Therefore, the claim holds in the limit as $T \to \infty$ for $r = 1/s$. 
The argument is similar if $\ell$ is monotonically non-increasing. In that case, we can show that whenever $y' > F_x^{-1}(s)$, we have $\ell(x)  \geq s \ell(x, y', a(x))$. Thus, whenever $\ell(x)   \leq s \ell(y, a, x)$, we have $y \leq F_x^{-1}(s)$ or $F_x(y) \leq s$. Because, $F_x$ is calibrated, we again have that
\begin{align*}
\frac{1}{T} \sum_{t=1}^T \mathbb{I} \{ \ell(x_t)   \leq s \ell(y_t, a_t, x_t) \} \leq \sum_{t=1}^T \mathbb{I} \{ F_{x_t}(y_t) < s \} = s + o(T),
\end{align*}
and the claim holds with $r = 1/s$. 
\end{proof}

Note that this implies the same result for a distribution calibrated model, since distribution calibration implies quantile calibration. 





\section{Algorithms for Online Calibration}
\label{sec:app_algorithms}

Here, we introduce algorithms that enforce calibration in an online setting. This task is challenging because the data distribution is the result of a sequential decision-making task. This distribution is therefore non-stationary: it is determined by our actions. 



\paragraph{Setup}
At each time step $t=1,2,...$ we observe a stream of datapoints comprised of features $x_t \in \mathbb R^d$. After $x_t$ is revealed, a base uncalibrated model (e.g., a Bayesian optimization model) produces a forecast; we represent this forecast via a quantile function $Q_t : [0,1] \to \mathbb{R}$
that targets a label $y_t \in \mathbb{R}$.
We assume that labels $y_t$ are bounded with $|y_t| < B$, where $B > 0$.
We also assume that $Q_t$ is strictly increasing and differentiable. 
Below, we may sometimes use the notation $Q(p)$ for $p \notin [0,1]$; 
in such cases, we use the convention that $Q(p) = - \infty$ for $p < 0$ and $Q(p) = \infty$ for $p > 1$.

The model $Q_t$ may produce miscalibrated outputs; we seek to compose $Q_t$ with a recalibrator $R_t : [0,1] \to [0,1]$ such that $Q_t \circ R_t$ is calibrated. After we choose $Q_t \circ R_t$, nature reveals a label $y_t \in \mathbb{R}$. 
Our goal is to select $R_t$ such that online quantile calibration (\ref{eqn:calibration1}) holds. Specifically, we use $o_t(y_t, p) = \mathbb{I}\{ y_t \leq Q_t(p) \}$ as an indicator of the binary outcome that $y_t$ falls below the $p$-th quantile. Our goal is to choose $R_t$ such that for all $p > 0$
\begin{equation}
    \frac{1}{T}\sum_{t=1}^T o_t(y_t, R_t(p)) - p \to 0 \text{ as $T \to \infty$}.
    \label{eqn:app_cal}
\end{equation}

Crucially, we want (\ref{eqn:app_cal}) to hold on any sequence of $(Q_t, x_t, y_t)$.

\paragraph{Optimization Problem}

Our algorithms construct $R_t$ via an optimization problem. Specifically, we will consider $R_t$ of the form
$$
R_t(p) = \arg\min_q \left[ \psi(q) - \sum_{s=1}^{t-1} \ell_p(y_s, q) \right],
$$
where $\ell_p : \mathbb{R} \times [0,1] \to \mathbb{R}_+$ is a loss function that we will define and $\psi(y) : \mathbb{R} \to \mathbb{R}_{+}$ is a regularizer (possibly equal to zero everywhere).
We choose the loss $\ell_p$ such that minimizing the average $\ell_p$ over the previously observed data yields an estimate of the $p$-th conditional quantile.
More specifically, we seek to define $\ell_p$ such that the probability $p$ is remapped to a probability $q$ for which 
the event $\{y_t \leq Q_t(q)\}$ is observed a fraction $p$ of the time.
Examples of suitable losses $\ell_p$ include the pinball loss---a generalization of the L1 loss motivated by conditional quantile estimation that we define below---as well as the weighted misclassification loss, which yields an algorithm analogous to that of \citep{kuleshov2018accurate}. 

We establish that $R_t$ obtained via the above construction yields calibrated forecasts through online optimization, a set of techniques that can provably minimize a loss function on distribution-free (possibly adversarial) data. We start by defining our algorithm for one quantile; then we use it to define a complete recalibrator $R_t$. 


\subsubsection{Recalibrating One Quantile}

First, consider the simpler problem of finding a $q_t \in [0,1]$ such that $Q_t(q_t)$ is an estimate of the $p$-th conditional quantile, i.e., 
$
\frac{1}{T}\sum_{t=1}^T o_t(y_t, q_t) - p \to 0 \text{ as $T \to \infty$}.
$

\paragraph{The Pinball Loss}
Our strategy for computing $q_t$ relies on online optimization. Specifically, we define a loss $\ell(y,y')$ and an update rule for $q_t$ such that $\sum_{t=1}^T \ell(y_t,Q_t(q_t))$ is minimized. 
Our loss will be inspired by the pinball loss. Given a target quantile $p$, 
the pinball loss $\ell_p$ defined as
\begin{align*}
    \ell_p(y_t, y) 
    & = (y_t - y) \cdot p \cdot \mathbb{I}\{ y_t > y \} + (y-y_t) \cdot (1-p) \cdot \mathbb{I}\{ y_t \leq y \} \\
    & = (y - y_t)(o_t(y_t, y) - p).
\end{align*}
Observe that $\ell_p$ is convex: its graph is V-shaped with the slopes of the two lines defining the V being $p$ and $1-p$. When $p=0.5$, the pinball loss coincides with the L1 loss (up to a multiplicative scaling factor). The pinball loss $\ell_p$ is interesting because the minimizer of $\ell_p$ over a set of datapoints $y_t$ yields a consistent estimator for the $p$-th quantile of this set of datapoints. 

Our algorithm optimizes a modification of the pinball loss which we call the quantile pinball loss (QPL) and which is defined as
\begin{align*}
    \ell_{tp}(y_t, q) 
    & = (Q_t^{-1}(y_t) - q) \cdot p \cdot \mathbb{I}\{ y_t > Q_t(q) \} + (q-Q_t^{-1}(y_t)) \cdot (1-p) \cdot \mathbb{I}\{ y_t \leq Q_t(q) \} \\
    & = (q-Q_t^{-1}(y_t))(o_t(y_t, q) - p).
\end{align*}

Note that the QPL still has the same V-shape as the original quantile loss, with each part of the V having a slope of $p$ and $(1-p)$. We can also show that the QPL features the same attractive property as the pinball loss in that it serves as a quantile estimator.

\begin{lemma} 
    The quantile pinball loss serves as a quantile estimator, in that $\arg\min_q  \sum_{s=1}^t \ell_{sp}(y_s, q)$ over a dataset $(y_s)_{s=1}^T$ yields a $p$-th quantile of the dataset.
\end{lemma}
\begin{proof}
        Note that the QPL is convex, as it is the weighted sum of two convex functions, $(q - Q_t^{-1}(y_t))_+$ and $(q)_+$. We minimize the QPL by setting its derivative to zero, giving:
    \begin{align*}
        0 &= \frac{d}{dq} \sum_{s=1}^t \ell_{sp}(y_s, q) \\
        &= \frac{d}{dq} \sum_{s=1}^t (q - Q_t^{-1}(y_t))(o_t(y_t, q) - p) \\
        &= \sum_{s=1}^t (o_t(y_t, q) - p) \\
    \end{align*}
    Thus, the minimum is achieved by a $q$ in the $p$-th quantile of $(y_s)_{s=1}^T$.
\end{proof}

\paragraph{Regularized Online Gradient Descent}

We consider the online optimization problem where at each step we choose a prediction $q_t$ for the $p$-th quantile. Nature then reveals $y_t$ and we incur the quantile pinball loss $\ell_p(y_t, q_t)$. 
We optimize this problem via regularized online gradient descent (OGD). Recall that OGD is an online optimization method for optimizing a sequence of functions. Note that OGD on the QPL is equivalent to Follow-The-Regularized-Leader (FTLR) on the linearized QPL. Thus, we introduce $\ell_t(q)$ as the linearization of the QPL at $q=q_t$, defined as 
\begin{align*}
    \ell_{t}(q) := (q - q_t) \partial_q \ell_{tp}(y_t, q_t) = (q - q_t) (o_t(y_t, q_t) - p)
\end{align*}
where the constant does not depend on $q$. The linearization $\ell_{t}(q)$ approximates $\ell(y_t, q)$ everywhere by the supporting hyperplane given by a subgradient at $q=q_t$. Then, at step $t$, we choose $q_t$ to minimize
\begin{equation}
    q_t = \arg\min_q ( \psi(q) + \sum_{s=1}^{t-1} \ell_s(q)), \label{eqn:app_ftrl}
\end{equation}
where $\psi(q) : \mathbb{R} \to \mathbb{R}_{+}$ is a regularizer. Observe that if we choose $\psi(q) = \frac{1}{2\eta} q^2$, this choice yields an exact solution to (\ref{eqn:app_ftrl}), where $q_t = \sum_{s=1}^{t-1} \eta g_s$ and $g_s \in \partial \ell_s(q_s)$ is a subgradient at the points $q_s$ for $s < t$. This derivation shows that we can compute the set of $q_t$ via gradient descent (although this is not strictly necessary). 

\paragraph{Quantile Calibration}
Online gradient descent normally yields guarantees on the regret of a model. Here, we also show that minimizing the quantile pinball loss induces quantile calibration. This is a condition that is derived from the average gradient of the function, as opposed to the regret. 
%
We first establish a technical lemma, then use the lemma to establish quantile calibration. The arguments for Lemma \ref{lem:bounded_quantile} and Theorem \ref{thm:consistentsingle} are inspired by results shown for adaptive conformal inference \citep{gibbs2021adaptive}.

\begin{lemma}
\label{lem:bounded_quantile}
    For any $t$, we have that $q_t$ is contained in $[-\eta, 1+\eta]$.
\end{lemma}

\begin{proof}
 Suppose not, and let $t$ be the first time step for which $q_t < - \eta$ (the case for $q_t > 1 + \eta$ is identical). Note that $|q_t - q_{t - 1}| = \eta | o_{t-1}(y_{t-1}, q_{t - 1}) - p| \leq \eta$. Thus, we know that $q_t < q_{t - 1} < 0$. The first inequality comes from the minimality of $t$, and the second comes from the fact that $|q_t - q_{t - 1}| \leq \eta$. However, $q_{t-1} < 0$ implies that $Q_{t-1}(q_{t-1}) = - \infty$. Thus, $o_t(y_t, q_t) = 1$ and $q_t = q_{t - 1} + \eta (o_t(y_t, q_t) - p) > q_{t - 1}$. This contradicts that $q_t < q_{t - 1}$.
\end{proof}

\consistentsingle*

\begin{proof}
    Observe that the subgradient $g_t \in \partial \ell_t(q_t)$  at the point $q_t$ can be written as $g_t = o_t(y_t, q_t) - p$. Thus, we can write
    \begin{align*}
        q_t = \sum_{s=1}^{t-1} \eta g_s = \sum_{s=1}^{t-1} \eta(o_s(y_s, q_s) - p).
    \end{align*}
    Dividing both sides by $t \eta $ gives
    \begin{align*}
        \frac{q_t}{t \eta} = \frac{1}{t}\sum_{s=1}^t(o_s(y_s, q_s) - p).
    \end{align*}
    Taking the absolute value and applying the lemma that $q_t \in [-\eta, 1 + \eta]$ gives the desired result. 
    
\end{proof}

This proves that the above algorithm yields a valid method for one quantile $p$.

\subsubsection{Quantile Function Recalibration}

Consider now a setting where we seek to define a full recalibrator $R(p)$. We define the recalibrator $R(p)$ for each $p$ via the algorithm in the previous section. Moreover, we can compute an approximate $R(p)$ by computing it at a fixed number of quantiles, and then interpolating. Alternatively, we may compute $R(p)$ at an arbitrary $p$ by solving the optimization problem.



    


\consistencyall*

\begin{proof}
    The inequality is a direct application of Theorem \ref{thm:consistentsingle}, where $R_t(p)$ is the value $q$ that minimizes the FTRL objective \eqref{eqn:ftrl}. 
\end{proof}





\end{document}